\documentclass[10pt,journal,compsoc]{IEEEtran}
\usepackage{microtype}
\usepackage{graphicx}
\usepackage{subfigure}
\usepackage{bbm}
\usepackage{color}
\usepackage{multirow}
\usepackage{amsmath}
\usepackage{amsfonts}
\usepackage{booktabs}
\usepackage{hyperref}

\ifCLASSOPTIONcompsoc
  \usepackage[nocompress]{cite}
\else
  \usepackage{cite}
\fi
\ifCLASSINFOpdf
\else
\fi
\usepackage{algorithmic}
\usepackage{algorithm}
\usepackage{url}

\begin{document}
%
\title{A Hamiltonian Monte Carlo Method for Probabilistic Adversarial Attack and Learning}
%
%
%
%

\author{Hongjun Wang, Guanbin Li, Xiaobai Liu and Liang Lin

\thanks{This work was supported in part by the National Key Research and Development Program of China under Grant
No.2018YFC0830103, in part by the National Natural Science Foundation of China under Grant No.61976250, No.61702565 and No.U1811463, in part by the Guangdong Basic and Applied Basic Research Foundation under Grant No.2020B1515020048, in part by National High Level Talents Special Support Plan (Ten Thousand Talents Program). This work was also sponsored by CCF-Tencent Open Research Fund. (Corresponding author: Guanbin Li)}
\thanks{H. Wang, G. Li and L. Lin are with the school of Data and Computer Science, Sun Yat-sen University, Guangzhou 510006, China (e-mail: wanghq8@mail2.sysu.edu.cn; liguanbin@mail.sysu.edu.cn; linliang@ieee.org).}
\thanks{X. Liu is with the Department of Computer Science, San Diego State University, San Diego, CA, 92182, USA (e-mail: xiaobai.liu@sdsu.edu).}}

\markboth{IEEE TRANSACTIONS ON PATTERN ANALYSIS AND MACHINE INTELLIGENCE}%
{ }
%



\IEEEtitleabstractindextext{%
\begin{abstract}
Although deep convolutional neural networks (CNNs) have demonstrated remarkable performance on multiple computer vision tasks, researches on adversarial learning have shown that deep models are vulnerable to adversarial examples, which are crafted by adding visually imperceptible perturbations to the input images.
Most of the existing adversarial attack methods only create a single adversarial example for the input, which just gives a glimpse of the underlying data manifold of adversarial examples. 
An attractive solution is to explore the solution space of the adversarial examples and generate a diverse bunch of them, which could potentially improve the robustness of real-world systems and help prevent severe security threats and vulnerabilities.
In this paper, we present an effective method, called Hamiltonian Monte Carlo with Accumulated Momentum (HMCAM), aiming to generate a sequence of adversarial examples.
To improve the efficiency of HMC, we propose a new regime to automatically control the length of trajectories, which allows the algorithm to move with adaptive step sizes along the search direction at different positions.
Moreover, we revisit the reason for high computational cost of adversarial training under the view of MCMC and design a new generative method called Contrastive Adversarial Training (CAT), which approaches equilibrium distribution of adversarial examples with only few iterations by building from small modifications of the standard Contrastive Divergence (CD) and achieve a trade-off between efficiency and accuracy. 
Both quantitative and qualitative analysis on several natural image datasets and practical systems have confirmed the superiority of the proposed algorithm.
\end{abstract}

}

\maketitle

\IEEEdisplaynontitleabstractindextext

%
\IEEEpeerreviewmaketitle

\IEEEraisesectionheading{\section{Introduction}\label{sec:introduction}}

%
%
%
%
\IEEEPARstart{W}{ith} the rapid development and superior performance achieved in various vision tasks, deep convolutional neural networks (CNNs) have eventually led to pervasive and dominant applications in many industries. 
However, most deep CNN models could be easily misled by natural images with imperceptible but deceptive perturbations.
These crafted images are known as adversarial examples, which have become one of the biggest threats in real-world applications with security-sensitive purposes\cite{Sharif16AdvML,DBLP:journals/corr/abs-1802-06430,Wang_2020_CVPR}.
Devising an effective algorithm to generate such deceptive examples can not only help to evaluate the robustness of deep models, but also promote better understanding about deep learning for the future community development. 

In the past literature, most state-of-the-art methods are well-designed for generating a \emph{single} adversarial example only, for example, by maximizing the empirical risk minimization (ERM) over the target model, and might not be able to exhaustively explore the solution space of adversarial examples.
In our opinion, adversarial examples of a deep model might form an underlying data manifold\cite{DBLP:conf/iclr/GilmerMFSRWG18,DBLP:conf/iclr/SongKNEK18,DBLP:journals/corr/TanayG16,stutz2019disentangling} rather than scattered outliers of the classification surface.
\begin{figure}[!tb]
\centering
\includegraphics[scale=0.3]{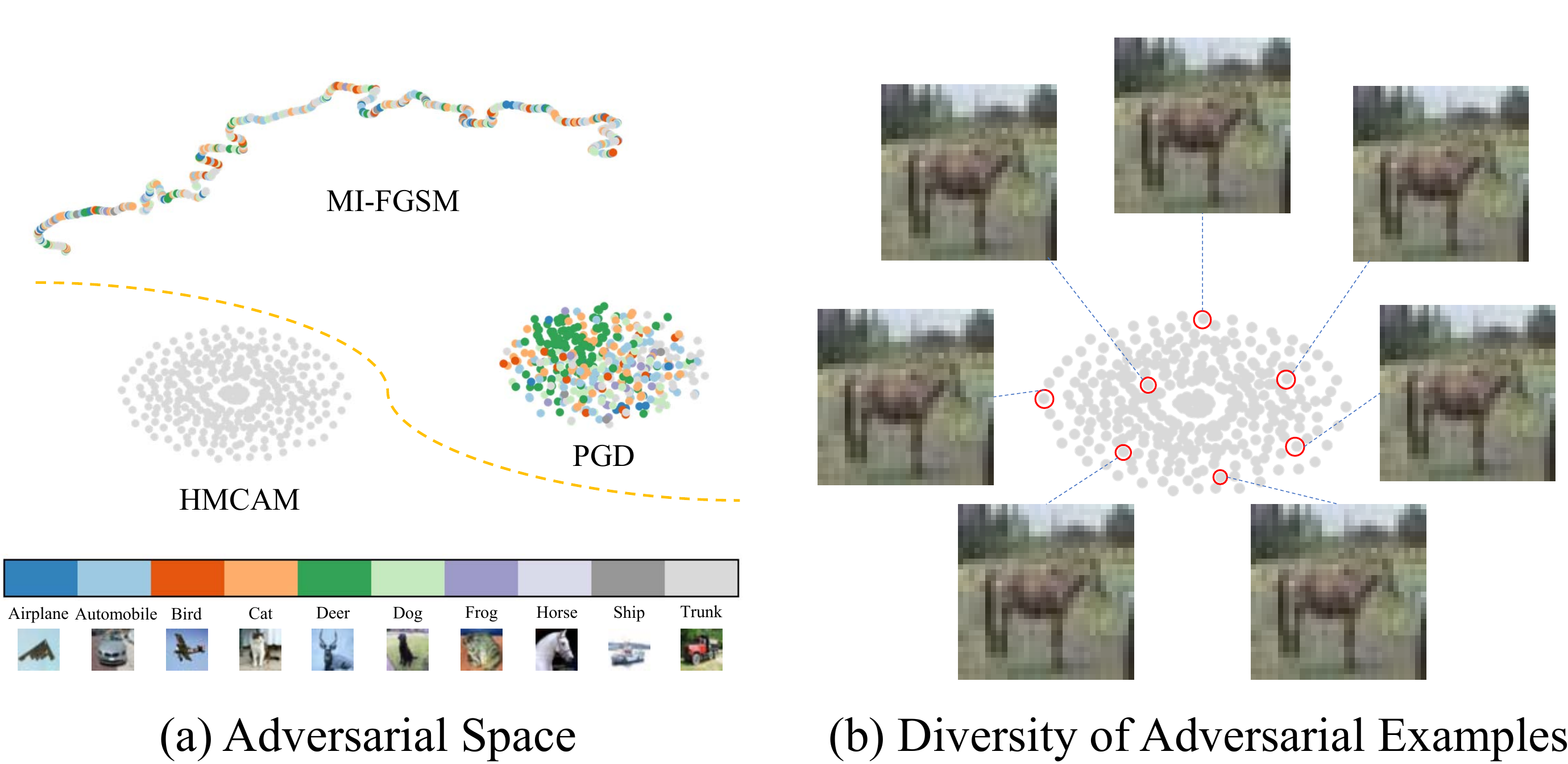}
\caption{Iterative Deterministic Generator vs. Stochastic MCMC-based Generator. We choose a natural image to generate 500 adversarial examples and visualize these samples by t-SNE\cite{maaten2008visualizing}. In contrast to two typical iterative deterministic methods (PGD\cite{DBLP:conf/iclr/MadryMSTV18} with 500 random restarts and MI-FGSM\cite{dong2018boosting} selecting samples at the final 500 iterations), MCMC-based method explores the solution space of adversarial examples and finds out the decision boundary of target classifier which is easily misled to erratic discrimination, then generates multiple diverse adversarial examples to attack. It is clear that our method automatically generates all of the 500 samples for a certain category in the untargeted attack scenario. When gradually increasing the number of MCMC sampling, the generated sequence of adversarial examples and their corresponding frequencies collectively depict the true underlying distribution of adversarial examples.}\label{fig:first}
\end{figure}
Therefore, we argue that it is desirable and critical for adversarial attack and learning methods to have the ability of generating multiple diverse adversarial examples in one run for the following reasons.
First, the diversity of adversarial examples can fully verify the robustness of an unknown system. 
Second, developing an attack with multiple distinct adversarial examples would enable adversarial training with such examples, which could make the model more robustness against white-box attacks.
Third, it is necessary to preserve multiple adversarial examples since the solution space of adversarial examples only depends on the targeted model and its input image even if the objective energy function of adversarial examples is constantly being improved\cite{bhagoji2017exploring,carlini2017towards,zheng2019distributionally,rony2019decoupling}, e.g. mapping the clipped gradient descent into tanh space or adding KL-divergence term. A series of adversarial samples can better depict the manifold of the solution space than a single global optimal, which can also bring more stable and superior performance on attacking.
In fact, training these representative generative models also suffers from instability due to the difficulty of finding the exact Nash equilibrium\cite{heusel2017gans,farnia2020gans} or tackling memorization\cite{DBLP:journals/corr/RadfordMC15,webster2019detecting,DBLP:conf/iclr/GulrajaniRM19}.

Motivated by the aforementioned observations, we rethink the generation of adversarial examples from the view of probabilistic distribution and develop an innovative paradigm called Hamiltonian Monte Carlo with Accumulated Momentum (HMCAM) for generating a sequence of adversarial examples in one run.
Given the attack objective energy function, the HMCAM method first constructs a joint distribution by Hamiltonian equations and the Metropolis-Hastings algorithm is used to determine whether to transition to the candidate sample via the acceptance function based upon the proposal distribution and the candidate-generating density. To improve the efficiency of HMC, we further propose a new regime called accumulated momentum to adaptively control the step sizes, which allows the algorithm to move with different step sizes along the search direction at different positions.
Conceptually, our HMCAM paradigm also reveals the roles of the well-known FSGM family algorithms, including FSGM\cite{DBLP:journals/corr/SzegedyZSBEGF13}, I-FGSM\cite{DBLP:conf/iclr/KurakinGB17}, PGD\cite{DBLP:conf/iclr/MadryMSTV18} and MI-FGSM\cite{dong2018boosting}. These methods can be considered as special cases of HMC with minor modifications. 
Inspired by our new paradigm, we further design a new generative method, called Contrastive Adversarial Training (CAT) , which approaches equilibrium distribution of adversarial examples with only few iterations by building from small modifications of the standard Contrastive Divergence\cite{hinton2002training}. 
We verify the effectiveness of both the adversarial attack and the training algorithms in multiple scenarios.
For the investigation of adversarial attack, we test our algorithm on single and ensemble models in both white-box and black-box manners. Extensive experiments conducted on the CIFAR10 dataset show that our method achieves much higher success rates with fewer iterations for black-box models and maintains similar success rates for white-box models. 
We also evaluate the proposed HMCAM on the CAAD 2018 defense champion solution\cite{xie2019feature}. It outperforms the official baseline attack and M-PGD (PGD with momentum) by a large margin, which clearly demonstrates the effectiveness of the proposed adversarial method.
To further show the practical applicability of our proposed method, we launch our attack on the real-world celebrity recognition system such as Clarifai, AWS and Azure. Compared with traditional iterative attack methods, HMCAM is able to generate more successful malicious examples to fool the systems through sampling from the likelihood models. 
For adversarial training, our CAT algorithm achieves much higher robustness than any other state-of-the-art adversarial training methods on both the CIFAR-10 and MNIST datasets and reaches a balance of performance and efficiency.
In summary, this paper has the following contributions:
\begin{itemize}
  \item We formulate the problem of generating adversarial examples in a HMC framework, which can produce multiple fair samples and better represent the underlying distribution of the adversarial examples. These fair samples can well reflect the typical state of the underlying system. 
  \item We design a new regime called accumulated momentum to adaptively control the step sizes, which allows the algorithm to move with different step sizes along the search direction at different positions, and thus improves the efficiency of HMC.
  \item We thoroughly compare the effectiveness of our algorithms in various settings against several iterative attack methods on both CIFAR10 and ImageNet, including the champion solution in the defense track of CAAD 2018 competitions. We also investigate the high efficiency of HMC framework in adversarial training and show the practical applicability of our HMCAM by successfully attacking the real-world celebrity recognition system.
\end{itemize}

\section{Related Work}
\textbf{Adversarial Attacks.} Since Szegedy \emph{et al.}\cite{DBLP:journals/corr/SzegedyZSBEGF13} first revealed that deep learning models were vulnerable to adversarial attacks, learning how to generate adversarial examples has quickly attracted wide research interest. Goodfellow \emph{et al.}\cite{DBLP:journals/corr/GoodfellowSS14} developed a single gradient step method to generate adversarial examples, which was known as the fast gradient sign method (FGSM). Kurakin \emph{et al.}\cite{DBLP:conf/iclr/KurakinGB17} extended FGSM to an iterative version and obtained much stronger adversarial examples. Based on their works, Madry \emph{et al.}\cite{DBLP:conf/iclr/MadryMSTV18} started projected gradient descent (PGD) from several random points in the $L_{\infty}$-ball around the natural example and iterate PGD. Dong \emph{et al.}\cite{dong2018boosting} proposed to add the momentum term into iterative process to boost adversarial attacks, which won the first places in the NIPS 2017 Adversarial Attacks and Defenses Competition. Due to the high efficiency and high success rates, the last two methods have been widely used as baseline attack models in many competitions. Our method also belongs to the iterative attack family but has much faster convergence and better transferability than alternative methods. When compared with recent similar works on distributional attack\cite{li2019nattack,zheng2019distributionally}, our HMC-based methods can better explore the distribution space of adversarial samples and reveal the reason for the high computational cost of adversarial training from the perspective of MCMC.
\textbf{Adversarial Defense.} To deal with the threat of adversarial examples, different strategies have been studied with the aim of finding countermeasures to protect ML models. 
These approaches can be roughly categorized into two main types: (a) detection only and (b) complete defense. The goal of the former approaches\cite{li2017adversarial,bhagoji2018enhancing,DBLP:conf/iclr/MetzenGFB17,DBLP:conf/iclr/Ma0WEWSSHB18,lee2018simple,tao2018attacks,zhang2018detecting} is to reject the potential malignant samples before feeding them to the ML models. However, it is meaningless to pinpoint the defects for developing more robust ML models. 
Complimentary to the previous defending techniques, the latter defense methods often involve modifications in the training process. For example, gradient masking\cite{papernot2016distillation,papernot2017extending,DBLP:conf/icml/AthalyeC018} or randomized models\cite{liu2018towards,DBLP:conf/iclr/XieWZRY18,10.1145/3357384.3357999,lecuyer2019certified,DBLP:conf/iclr/LiuLWH19} obfuscate the gradient information of the classifiers to confuse the attack mechanisms. There are also some add-on modules\cite{xie2019feature,akhtar2018defense,DBLP:journals/corr/GuR14,liao2018defense,li2019rosa,he2019non} being appended to the targeted network to protect deep networks against the adversarial attacks. Besides all the above methods, adversarial training\cite{DBLP:journals/corr/GoodfellowSS14,DBLP:conf/iclr/KurakinGB17,kannan2018adversarial,DBLP:conf/iclr/MadryMSTV18,liu2019robgan} is the most effective way, which has been widely verified in many works and competitions. However, limited works\cite{shafahi2019adversarial,zhang2019you} focus on boosting robust accuracy with reasonable training time consumption.

\textbf{Markov Chain Monte Carlo Methods.} Markov chain Monte Carlo (MCMC)\cite{neal1993probabilistic} established a powerful framework for drawing a series of fair samples from the target distribution. But MCMC is known for its slow convergence rate which prevents its wide use in time critical fields. To address this issue, Hamiltonian (or Hybrid) Monte Carlo method (HMC) \cite{duane1987hybrid,neal2011mcmc} was introduced to take advantages of the gradient information in the target solution space and accelerate the convergence to the target distribution. Multiple variants of HMC \cite{pasarica2010adaptively,salimans2015markov,hoffman2014no} were also developed to integrate adaptive strategies for tuning step size or iterations of leapfrog integrator.
Recently, the fusion of MCMC and machine learning hastens wide range of applications, including data-driven MCMC\cite{tu2002image,chen2014stochastic}, adversarial training\cite{song2017nice}, cooperative learning\cite{xie2018cooperative}, which shows great potential of MCMC in deep learning.

\section{Methodology}
In this section, we briefly review the Markov chain Monte Carlo (MCMC) method \cite{neal1993probabilistic} and Hamiltonian Monte Carlo (HMC) methods \cite{duane1987hybrid,neal2011mcmc}. Then we will explain that most of the existing methods for generating adversarial examples are the specializations of HMC. Finally, we illustrate how to modify the update policy of the momentum item in HMC to obtain a better trajectory.

\subsection{Review: MCMC and Hamiltonian Monte Carlo}
We now give the overall description of Metropolis-Hasting based MCMC algorithm. Suppose $p$ is our target distribution over a space $\mathcal{D}$, MCMC methods construct a Markov Chain that has the desired distribution $p$ as its stationary distribution.
At the first step, MCMC chooses an arbitrary point $x_0$ as the initial state. Then it repeatedly performs the dynamic process consisting of the following steps: (1) Generate a candidate sample $\tilde{x}$ as a ``proposed'' value for $x_{t+1}$ from the candidate-generating density $Q(x_t|\tilde{x})$, which generates a value $\tilde{x}$ from $Q(x_t|\tilde{x})$ when a process is at the state $x_t$. (2) Compute the acceptance probability $\xi=\min (1, \frac{p(\tilde{x}) Q(x_{t} | \tilde{x})}{p(x_{t}) Q(\tilde{x} | x_{t})})$, which is used to decide whether to accept or reject the candidate. (3) Accept the candidate sample as the next state with probability $\xi$ by setting $x_{t+1}=\tilde{x}$. Otherwise reject the proposal and remain $x_{t+1}=x_t$.
Although MCMC makes it possible to sample from any desired distributions, its random-walk nature makes the Markov chain converge slowly to the stationary distribution $p(x)$. 
 
In contrast, HMC employs physics-driven dynamics to explore the target distribution, which is much more efficient than the alternative MCMC methods.
Before introducing HMC, we start out from an analogy of Hamiltonian systems in \cite{neal2011mcmc} as follows. Suppose a hockey puck sliding over a surface of varying height and both the puck and the surface are frictionless. 
The state of the puck is determined by \emph{potential energy} $U(\theta)$ and \emph{kinetic energy} $K(v)$, where $\theta$ and $v$ are the position and the momentum of the puck. The evolution equation is given by the Hamilton's equations:
\begin{equation}
\label{eqn:evolution}
\left\{
\begin{array}{l}
{\frac{\partial \theta}{\partial t}=\frac{\partial H}{\partial v}=\nabla_v K(v)} \\ 
{\frac{\partial v}{\partial t}=\frac{\partial H}{\partial \theta}=-\nabla_\theta U(\theta)}.
\end{array}\right.
\end{equation}
Due to the reversibility of Hamiltonian dynamics, the total energy of the system remains constant: 
\begin{equation}
\label{eqn:total}
H(\theta,v)=U(\theta)+K(v).
\end{equation}

\begin{algorithm}[!tb]
    \centering
    \caption{Hamiltonian Monte Carlo}
    \label{alg:HMC}
    \begin{flushleft}
      \textbf{Inputs:} Target distribution $p(\theta)$, initial position $\theta^{(1)}$ and step size $\alpha$ \\
    \end{flushleft}
    \begin{algorithmic}[1]
      \STATE \emph{/*Hamiltonian system construction*/}
      \STATE $U(\theta)=-\log p(\theta)$, $K(v)=v^{T} \mathbf{I}^{-1} v / 2$
      \FOR{$s=1,2,\cdots$}  
        \STATE $v_0 \sim \mathcal{N}(0, \mathbf{I})$, $\theta_0=\theta^{(s)}$
        \STATE \emph{/*Leapfrog integration*/}
        \STATE $v_{0} \leftarrow v_{0}-\frac{\alpha}{2} \nabla U\left(\theta_{0}\right)$
        \FOR{$t=1$ to $T$}  
          \STATE $\theta_{t} \leftarrow \theta_{t-1}+\alpha \nabla K\left(v_{t-1}\right)$
          \STATE $v_{t} \leftarrow v_{t-1}-\alpha \nabla U\left(\theta_{t}\right)$
        \ENDFOR
        \STATE $v_{T} \leftarrow v_{T}-\frac{\alpha}{2} \nabla U\left(\theta_{T}\right)$
        \STATE \emph{/*Metropolis-Hastings correction*/}
        \STATE $u \sim$ Uniform(0,1)
        \IF{$u<\operatorname{min}(1,e^{H(\theta_T,v_T)-H(\theta_s,v_s)})$}
          \STATE $\theta^{(s+1)}\leftarrow \theta_T$
        \ELSE
          \STATE $\theta^{(s+1)}\leftarrow \theta^{(s)}$
        \ENDIF
      \ENDFOR
    \end{algorithmic}
\end{algorithm}
As for HMC, it contains three major parts: (1) Hamiltonian system construction; (2) Leapfrog integration; (3) Metropolis-Hastings correction. Firstly, the Hamiltonian is an energy function for the joint density of the variables of interest $\theta$ and auxiliary momentum variable $v$, so HMC defines a joint distribution via the concept of a canonical distribution:
\begin{equation}
\label{eqn:joint}
p(\theta, v)\propto \exp \left(\frac{-H(\theta,v)}{\tau}\right),
\end{equation}
where $\tau=1$ for the common setting.
Then, HMC discretizes the system and approximately simulates Eq. (\ref{eqn:evolution}) over time via the leapfrog integrator. 
Finally, because of inaccuracies caused by the discretization, HMC performs Metropolis-Hastings\cite{metropolis1953equation} correction without reducing the acceptance rate. A full procedure of HMC is described in Algorithm \ref{alg:HMC}. 

According to Eq. (\ref{eqn:total}) and (\ref{eqn:joint}), the joint distribution can be divided into two parts:
\begin{equation}
\label{eqn:HMC_part}
p(\theta, v)\propto \exp \left(\frac{-U(\theta)}{\tau}\right) \exp \left(\frac{-K(v)}{\tau}\right).
\end{equation}
Since $K(v)$ is an auxiliary term and always setting $K(v)=v^{T} \mathbf{I}^{-1} v / 2$ with identity matrix $\mathbf{I}$ for standard HMC, our aim is that the potential energy $U(\theta)$ can be defined as $U(\theta) = -\log p(\theta)$ to explore the target density $p$ more efficiently than using a proposal probability distribution. If we can calculate $\nabla_\theta U(\theta)=-\frac{\partial \log(p(\theta)) }{\partial \theta}$, then we can simulate Hamiltonian dynamics that can be used in an MCMC technique.

\subsection{Simulating Adversarial Examples Generating by HMC}\label{sec_HMC}
Considering a common classification task, we have a dataset $\mathcal{D}$ that contains normalized data $x \in[0,1]^{d}$ and their one-hot labels $y$. We identify a target DNN model with an hypothesis $f(\cdot)$ from a space $\mathcal{F}$. The cross entropy loss $J$ function is used to train the model. Assume that the adversarial examples for $x$ with label $y$ are distributed over the solution space $\Omega$. Given any input pair $(x,y)$, for a specified model $f(\cdot)\in \mathcal{F}$ with fixed parameters, the adversary wants to find such examples $\tilde{x}$ that can mislead the model:
\begin{equation}
\Omega =\arg \max_{N(x) \subset \mathcal{N}(x)} \int J\left(\tilde{x},y\right) p\left(\tilde{x} | x, y\right) d \tilde{x},
\end{equation}
where $\mathcal{N}(x)$ is the neighboring regions of $x$ and defined as $x^{\prime} \in \mathcal{N}(x):=\left\{\left\|x^{\prime}-x\right\|_{1,2,\operatorname{or}\infty} \leq \epsilon\right\}$. 
From the perspective of Bayesian statistics, we can make inference about adversarial examples over a solution space $\Omega$ from the posterior distribution of $\tilde{x}$ given the natural inputs $x$ and labels $y$. 
%
%
\begin{table*}[!htb]
\centering
\small
\begin{tabular}{l|ccc|cc|c}
\multirow{2}{*}{Methods}                                      & \multicolumn{3}{c|}{Hamiltonian system construction}                                                                           & \multicolumn{2}{c|}{Iteration}                        & \multirow{2}{*}{\begin{tabular}[c]{@{}c@{}}Metropolis-Hastings \\ correction\end{tabular}} \\ \cline{2-6}
                                                              & potential energy?                       & kinetic energy?                         & sampling?                                  & $\theta$ update?          & $v$ update?               &                                                                                               \\ \hline\hline
FGSM\cite{DBLP:journals/corr/GoodfellowSS14} & \checkmark, but implicit & \checkmark, but implicit & $\mathrm{x}$                               & $\mathrm{x}$              & $\mathrm{x}$              & $\mathrm{x}$                                                                                  \\
I-FGSM\cite{DBLP:conf/iclr/KurakinGB17}      & \checkmark, but implicit & \checkmark, but implicit & $\mathrm{x}$                               & \checkmark & $\mathrm{x}$              & $\mathrm{x}$                                                                                  \\
PGD\cite{DBLP:conf/iclr/MadryMSTV18}         & \checkmark, but implicit & \checkmark, but implicit & \checkmark, but independent & \checkmark & $\mathrm{x}$              & $\mathrm{x}$                                                                                  \\
MI-FGSM\cite{dong2018boosting}               & \checkmark, but implicit & \checkmark, but implicit & $\mathrm{x}$                               & \checkmark & \checkmark & $\mathrm{x}$                                                                                 
\end{tabular}
\caption{Relationship between HMC and the family of fast gradient sign methods.}\label{tab:HCM_FGSM}
\end{table*}
\begin{equation}
\label{eqn:posterior}
\tilde{x} \sim p(\tilde{x} | x, y) \propto p(y | \tilde{x})p(\tilde{x} | x), \quad \tilde{x} \in \Omega.
\end{equation}
%
In Hamiltonian system, it becomes to generate samples from the joint distribution $p(\theta, v)$.
Let $\theta=\tilde{x}$, according to Eq. (\ref{eqn:posterior}) and (\ref{eqn:HMC_part}), we can express the posterior distribution as a canonical distribution (with $\tau=1$) using a potential energy function defined as:
\begin{equation}
\begin{aligned} 
U                     &=\frac{1}{N} \sum_{i=1}^{N} -\log p(y^{(i)} | \tilde{x}^{(i)})-\log p(\tilde{x} | x) \\ 
                      &= J\left(\tilde{x},y\right)-\log p(\tilde{x} | x).
\end{aligned}
\end{equation}
Since $J\left(\tilde{x},y\right)$ is the usual classification likelihood measure, the question remains how to define $p(\tilde{x}|x)$. A sensible choice is a uniform distribution over the $L_p$ ball around $x$, which means we can directly use a DNN classifier to construct a Hamiltonian system for adversarial examples generating as the base step of HMC. 

Recall that the development of adversarial attacks is mainly based on the improvement of the vanilla fast gradient sign method, which derives I-FGSM, PGD and MI-FGSM. For clarity, we omit some details about the correction due to the constraint of adversarial examples. The core policy of the family of fast gradient sign methods is:
\begin{equation}
\label{eqn:FSGM_core}
\tilde{x}_{t}=\tilde{x}_{t-1}+\alpha \cdot \operatorname{sign}(g_{t}),
\end{equation}
where $g_t$ is the gradient of $J$ at the $t$-th iteration, i.e., $\nabla_{x}J(\tilde{x}_{t-1}, y)$. It is clear that the above methods are the specialization of HMC by setting:
\begin{equation}
\begin{aligned} 
\label{eqn:setting}
\theta_t=\tilde{x}_t&, \quad v_t=g_t \\
H(\theta,v)=&J(\theta)+|v|.
\end{aligned}
\end{equation}
More specifically, I-FGSM can be considered as the degeneration of HMC, which explicitly updates the position item $\theta$ but implicitly changes the momentum item $v$ at every iteration. One of the derivation of I-FGSM, MI-FGSM, has explicitly updated both $\theta$ and $v$ by introducing $g_{t}=\mu g_{t-1}+\frac{1}{||\nabla J(\tilde{x}_{t-1},y)||_1}\nabla J(\tilde{x}_{t-1},y)$ after Eq. (\ref{eqn:FSGM_core}) at each step with the decay factor $\mu=1$. The other derivative PGD runs Eq. (\ref{eqn:FSGM_core}) on a set of initial points $\tilde{x}_{0}\in \left\{\tilde{x}_{0}^{(1)},\tilde{x}_{0}^{(2)},\cdots,\tilde{x}_{0}^{(S)}\right\}$ adding different noises, which can be treated as a parallel HMC but the results are mutually independent.

\begin{algorithm}[!tb]
    \centering
    \caption{HMCAM}
    \label{alg:ours}
    \begin{flushleft}
      \textbf{Inputs:} Target DNN model $f(\cdot)$ with loss function $J$, initial position $\theta^{(0)}=x$, step size $\alpha$, sampling transition $S$, updating iteration $T$, magnitude of perturbation $\varepsilon$ and small constant $\delta$ \\
      \textbf{Inputs:} exponential decay rates for the moment estimates $\beta_1=0.95$, $\beta_2=0.999$ \\
    \end{flushleft}
    \begin{algorithmic}[1]
        \STATE \emph{/*Hamiltonian system construction*/}
        \STATE $U(\theta)=J$, $K(v)=|v|$
        \FOR{$s=1$ to $S$}  
          \STATE Initialize $v_0 \leftarrow 0$; $e_0 \leftarrow 0$; $\hat{e}_{0} \leftarrow 0$
          \STATE \emph{/*Accumulated Momentum*/}
          \FOR{$t=1$ to $T$}  
            \STATE $b_{1} \leftarrow 1-\beta_{1}^{t}$, \quad $b_{2} \leftarrow 1-\beta_{2}^{t}$
            \STATE $v_{t} \leftarrow \beta_{1} \cdot v_{t-1}-\left(\beta_{1} - 1\right) \cdot \nabla_{\theta}J(\theta_{t-1}, y)$
            \STATE $e_{t} \leftarrow \beta_{2} \cdot e_{t-1}-\left(\beta_{2} - 1\right) \cdot \nabla_{\theta}^{2}J(\theta_{t-1}, y)$
            \STATE $\hat{e}_{t} \leftarrow \operatorname{max}(e_{t}, \hat{e}_{t})$
            \STATE $\theta_{t}\leftarrow \theta_{t-1}+\operatorname{min}(\varepsilon,\frac{\alpha}{b_{1}(\sqrt{\frac{\hat{e}_{t}}{b_{2}}}+\delta)}) \nabla_v K(v_t)$
            \STATE $\theta_{t}\leftarrow \mathbb{P}_{\operatorname{\varepsilon-ball}}(\theta_{t})$
          \ENDFOR
          \STATE \emph{/*Metropolis-Hastings correction*/}
          \STATE $u \sim$ Uniform(0,1)
          \IF{$u<\operatorname{min}(1,e^{H(\theta_T,v_T)-H(\theta_s,v_s)})$}
            \STATE $\theta^{(s+1)}\leftarrow \theta_T$
          \ELSE
            \STATE $\theta^{(s+1)}\leftarrow \theta^{(s)}$
          \ENDIF
        \ENDFOR
        \STATE \textbf{Return}  A sequence of adversarial examples $\left\{\theta\right\}$
    \end{algorithmic}
\end{algorithm}
\subsection{Adaptively Exploring the Solution Space with Accumulated Momentum}
Although the above formulation has proved that HMC can be used to simulate adversarial examples generating, one major problem of these methods is that $\theta$ and $v$ are not independent because of $v_t=\nabla J(\theta_{t-1})$ as discussed in Eq. (\ref{eqn:setting}). The other disadvantage is in optimization: SGD scales the gradient uniformly in all directions, which can be particularly detrimental for ill-scaled problems. Like the need to choose step size in HMC, the laborious learning rate tuning is also troublesome.

To overcome the above two problems, we present a Hamiltonian Monte Carlo with Accumulated Momentum (HMCAM) for adversarial examples generating. The resulting HMCAM algorithm is shown in Algorithm \ref{alg:ours}.
The core of our accumulated momentum strategy is using exponential moving average (EMA) to approximate the first and second moment of the stochastic gradient by weighted accumulating the history moment information. Let us initialize the exponential moving average as $v_0=e_0=0$. After $t$ inner-loop steps, the accumulated momentum $v_t$ is:
\begin{equation}
\begin{aligned} 
\label{eqn:EMA}
v_{t} &=\left(1-\beta_{1}\right) \sum_{i=1}^{t} \beta_{1}^{t-i} \nabla_{\theta}J(\theta_{i-1})\\
      &=(1-\beta_{1})\underbrace{\nabla_{\theta}J(\theta_{t-1})}_{\hbox{\footnotesize{Current term}}} \\ 
      &+\beta_{1}\underbrace{[\nabla_{\theta}J(\theta_{t-2})+\beta_1\nabla_{\theta}J(\theta_{t-3})+\cdots+\beta_1^{t-2}\nabla_{\theta}J(\theta_{0})]}_{\hbox{\footnotesize{History term}}}.
\end{aligned}
\end{equation}
The derivation for the second moment estimate $e_t$ is completely analogous. Owing to the fact that the decay rates $\beta_1$ close to 1 is typically recommended in practice, the contribution of older gradients decreases exponentially. But meanwhile, we can observe in Eq. (\ref{eqn:EMA}) that the current gradient only accounts for $1-\beta_1\rightarrow 0$, which is much smaller than $\beta_1$. This indicates that performing exponential moving averages for the step in lieu of the gradient greatly reduces the relevance between $v_t$ and the current position $\theta_{t-1}$. That makes the sequence of samples into an approximate Markov chain.

As for step size, there always be a tradeoff between using long trajectories to make HMC more efficient or using shorter trajectories to update more frequently. Ignoring small constant $\delta$, our accumulated momentum is to update the position by:
\begin{equation}
\begin{aligned} 
\label{eqn:update}
\theta_{t} &=\theta_{t-1}+\alpha\Delta\theta=\theta_{t-1}+\frac{\sqrt{1-\beta_2^t}}{(1-\beta_1^t)}\cdot\frac{\alpha}{|v_t|}\cdot\frac{v_t}{\sqrt{e_t}},
\end{aligned}
\end{equation}
where $\sqrt{1-\beta_2^t}/(1-\beta_1^t)$ corrects the biasd estimation of moments towards initial values at early stages due to the property of EMA.
When approaching to the minima, $\alpha/|v_t|$ automatically decreases the size of the gradient steps along different coordinates. Because $v_t/\sqrt{e_t}$ leads
to smaller effective steps in solution space when closer to zero, this anisotropic scale of step size helps $\theta$ to escape sharp local minimal at the later period of the learning process at some coordinates, which leads to better generalization. We apply similar idea as \cite{DBLP:conf/iclr/ReddiKK18} by replacing $e$ to $\hat{e}$ that maintains the maximum of all history $e$ to keep a non-increasing step size $(\sqrt{e_{t+1}}-\sqrt{e_{t}})/\alpha_{t}\succeq0$. To guarantee the step size does not exceed the magnitude of adversarial perturbations, we confines the $\alpha$ to a predefined maximum $\varepsilon$ by applying element-wise $\operatorname{min}$.
\begin{algorithm}[tb]
   \caption{Contrastive Adversarial Training}
   \label{alg:CAT}
\begin{algorithmic}
   \STATE {\bfseries Input:} A DNN classifier $f_{\omega}(\cdot)$ with initial learnable parameters $\omega_0$; training data $x$ with visible label $y$; number of epochs $N$; length of trajectory $K$; repeat time $T$; magnitude of perturbation $\varepsilon$; learning rate $\kappa$; step size $\alpha$.
   \STATE \emph{/*Stage-0: Construct Hamiltonian system*/}
   \STATE $U(\theta, \omega, \tilde\omega, y, k) = -J_{cd}\left(f_{\omega}(\theta^{k-1}), f_{\tilde\omega}(\theta^K), y\right)$, $\mathcal{K}(v) = |v|$
   \STATE Initialize $\omega=\tilde\omega=\omega_0$, $\theta^K=\theta^0$.
   \FOR {epoch$=1\cdots N/(TK)$}
   \STATE $\theta^0\leftarrow x+v_0$, $v_0\sim \operatorname{Uniform}(-\varepsilon,\varepsilon)$.
   \FOR{$t=1$ {\bfseries to} $T$}
   \STATE \emph{/*Stage-1: Generate adversarial examples by K-step contrastive divergence*/}
   \FOR{$k=1$ {\bfseries to} $K$}
   \STATE $\theta^{k} \leftarrow \theta^{k-1} + \varepsilon \cdot \nabla \mathcal{K}(v_{t-1})$
   \STATE $v_t \leftarrow v_{t-1} - \alpha\nabla U(\theta, \omega, \tilde\omega, y, k)$
   \STATE $v_t \leftarrow \operatorname{clip}(v_t, -\varepsilon, \varepsilon)$
   \ENDFOR
   \STATE \emph{/*Stage-2: Update parameters of DNN by generated adversarial examples*/}
   \STATE $\boldsymbol{g}_{\omega} \leftarrow \mathbb{E}_{(\theta, y)}\left[\nabla_{\omega} J_{ce}(f_{\omega}(\theta^{K}), y)\right]$
   \STATE $\tilde\omega \leftarrow \omega$
   \STATE $\omega \leftarrow \omega-\kappa \boldsymbol{g}_{\omega}$
   \ENDFOR
   \ENDFOR
\end{algorithmic}
\end{algorithm}

After every full inner iteration, we calculate the acceptance rate of the candidate sample by M-H sampling and reinitialize the first/second moment as well as the maximum of second moment to zero and then perform the next generation. 
M-H algorithm distributes the generating samples to staying in high-density regions of the candidate distribution or only occasionally visiting low-density regions through the acceptance probability. As more and more sample are produced, the distribution of samples more closely approximates the desired distribution and its returning samples are more in line with such distribution than other works like PGD with random starts.

\section{Contrastive Adversarial Training}
Assume softmax is employed for the output layer of the model $f(\cdot)$ and let $f(x)$ denote the softmax output of a given input $x \in \mathbb{R}^{d}$, i.e., $f(x):\mathbb{R}^{d} \rightarrow \mathbb{R}^{C}$, where $C$ is the number of categories. We also assume that there exists an oracle mapping function $f^*\in \mathcal{F}: x \mapsto y^*$, which pinpoints the belonging of the input $x$ to all the categories by accurate confidence scores $y^*\in \mathbb{R}^{C}$. The common training is to minimize the cross-entropy (CE) loss, which is defined as:
\begin{equation}
\label{eqn:regular_training}
f=\underset{f \in \mathcal{F}}{\arg \min } \quad \mathbb{E}_{\left(x, y\right) \sim \mathcal{D}}\left[L_{ce}\left(f(x), y\right)\right],
\end{equation}
where $y$ is the manual one-hot annotation of the input $x$ since $y^*$ is invisible. The goal of Eq. (\ref{eqn:regular_training}) is to update the parameters of $f$ for better approaching $f^*$, which leads to:
\begin{equation}
f(x)\approx y\approx y^*=f^*(x).
\end{equation}

Suppose the target DNN model correctly classifies most of the input after hundreds of iterations, it will still be badly misclassified by adversarial examples (i.e., $\arg \max _{c \in \left\{1,\cdots,C\right\}} f(\tilde{x})_c\neq y[c]$). 
In adversarial training, these constructed adversarial examples are used to updates the model using minibatch SGD.
The objective of this minmax game can be formulated as a robust optimization following\cite{DBLP:conf/iclr/MadryMSTV18}:
\begin{equation}
\label{eqn:adv_training}
f^{\prime}=\underset{f \in \mathcal{F}}{\arg \min } \underset{{\left(x, y\right) \sim \mathcal{D}}}{\mathbb{E}}\left[\max _{\tilde{x} \in \mathcal{N}(x)} L_{ce}\left(f\left(\tilde{x}\right), y\right)\right].
\end{equation}

As mentioned in Section \ref{sec_HMC}, the inner maximization problem can be reformulated as the process of HMC. It is obvious that the high time consumption of adversarial training is caused by the long trajectory of HMC. 
But running a full trajectory for many steps is too inefficient since the model changes very slightly between parameter updates. 
Thus, we take advantage of that by initializing a HMC at the state in which it ended for the previous model. 
This initialization is often fairly close to the model distribution, even though the model has changed a bit in the parameter update. Besides, the high acceptance rate of HMC indicates that it is not neccesary to run a long Markov Chain from the initial point. Therefore, we can simply run the chain for one full step and then update the parameters to reduce the tendency of the chain to wander away from the initial distribution on the first step instead of running the full trajectory to equilibrium.
We takes small number $K$ of transitions from the data sample $\left\{x_i\right\}^n_i=1$ as the initial values of the MCMC chains and then use these $K$-step MCMC samples to approximate the gradient for updating the parameters of the model. Algorithm\ref{alg:CAT} summarizes the full algorithm. 

Moreover, we also present a new training objective function $J_{cd}$, which minimizes the difference of KL divergence between two adjacent sampling steps to substitute the common KL loss:
\begin{equation}
\label{eqn:loss_cd}
J_{cd}=\rho (Q^{0}\left\|Q^{\infty}) - \lambda (Q^{1}\right\| Q^{\infty}),
\end{equation}
where $||$ denotes a Kullback-Leibler divergence and $\rho$ and $\lambda$ are the balanced factors. The intuitive motivation for using this $J_{cd}$ is that we would like every state in HMC exploring to leave the initial distribution $Q_0$ and $Q^{0}||Q^{\infty}$ would never exceed $Q^{1}||Q^{\infty}$ until $Q_1$ achieves the equilibrium distribution. We set $\lambda=2,\rho=1$ and analyze how this objective function influences the partial derivative of the output probability vector with respect to the input. Due to the fact that the equilibrium distribution $Q^{\infty}$ is considered as a fixed distribution and the chain rule, we only need to focus on the derivative of the softmax output vector with respect to its input vector in the last layer as follows:
\begin{equation}
\begin{aligned}
\label{eqn:dev_CD}
\nabla U_{\operatorname{last}} &=2\sum_{c} y_{c} \frac{\partial \log f_\omega(\tilde{x}^K)_{c}}{\partial \tilde{x}^{\prime}} - \sum_{c} y_{c} \frac{\partial \log f_{\tilde\omega}\left(\tilde{x}\right)_{c}}{\partial \tilde{x}^{\prime}}\\
                                        &=2f_{\omega}(\tilde{x}^K)_{c} \sum_{c} y_{c} - f_{\tilde\omega}\left(\tilde{x}\right)_{c} \sum_{c} y_{c} - y \\
                                        &=f_{\omega}(x^K) - (y - \Delta f), 
\end{aligned}
\end{equation}
where $\Delta f=f_{\omega}(x^K)-f_{\tilde\omega}\left(\tilde{x}\right)$. Based on this abbreviation, we can easily get the relationship between Eq. (\ref{eqn:dev_CD}) and $\frac{\partial J_{ce}}{\partial \tilde{x}^{\prime}}=f_{\omega}(x^K) - y$. For each adversarial example generation, Eq. (\ref{eqn:dev_CD}) makes an amendment of $y$ which is determined by the difference of current and the last $K$-step HMC samples output probability. Since $f_\omega$ and $f_\omega(x)$ are more closer to $f^*$ and $y^*$ than $f_{\tilde\omega}$ and $f_{\tilde\omega}(x)$, each update of $\tilde{x}$ would be better corrected.

\renewcommand{\arraystretch}{0.9}
\begin{table*}[!htb]
\centering
\begin{tabular}{l|lccccc}
Model                        & Attack         & ResNet32                                                               & VGG16                                                                  & ResNetXt                                                               & Densenet121                                                            & ResNet32\_A                       \\ \hline\hline
\multirow{5}{*}{ResNet32}    & FGSM           & \emph{38.31\%}                           & 29.30\%                                                                & 19.89\%                                                                & 22.64\%                                                                & 3.79\%                            \\
                             & PGD         & \emph{98.12\%}                           & 34.92\%                                                                & 49.44\%                                                                & 56.50\%                                                                & 4.60\%                            \\
                             & M-PGD        & \emph{\textbf{98.93\%}} & 37.89\%                                                                & 55.48\%                                                                & 61.01\%                                                                & 7.44\%                            \\
                             & AI-FGSM (Ours) & \emph{98.76\%}                           & 42.23\%                                                                & 58.12\%                                                                & 64.12\%                                                                & 9.56\%                            \\
                             & HMCAM (Ours)   & \emph{98.76\%}                           & \textbf{42.69\%}                                      & \textbf{58.76\%}                                      & \textbf{65.20\%}                                      & \textbf{10.01\%} \\ \hline\hline
\multirow{5}{*}{VGG16}       & FGSM           & 37.86\%                                                                & \emph{56.34\%}                           & 27.34\%                                                                & 31.54\%                                                                & 4.22\%                            \\
                             & PGD         & 59.39\%                                                                & \emph{80.55\%}                           & 50.50\%                                                                & 55.72\%                                                                & 5.52\%                            \\
                             & M-PGD        & 64.02\%                                                                & \emph{83.64\%}                           & 54.95\%                                                                & 60.48\%                                                                & 7.75\%                            \\
                             & AI-FGSM (Ours) & 64.77\%                                                                & \emph{86.76\%}                           & 53.38\%                                                                & 59.45\%                                                                & 9.83\%                            \\
                             & HMCAM (Ours)   & \textbf{68.60\%}                                      & \emph{\textbf{93.29\%}} & \textbf{55.39\%}                                      & \textbf{62.70\%}                                      & \textbf{10.26\%} \\ \hline\hline
\multirow{5}{*}{ResNetXt}    & FGSM           & 27.44\%                                                                & 28.52\%                                                                & \emph{31.74\%}                           & 24.03\%                                                                & 4.50\%                            \\
                             & PGD         & 65.48\%                                                                & 35.19\%                                                                & \emph{96.60\%}                           & 69.13\%                                                                & 6.55\%                            \\
                             & M-PGD        & 72.81\%                                                                & 38.50\%                                                                & \emph{\textbf{98.02\%}} & 76.55\%                                                                & 10.11\%                           \\
                             & AI-FGSM (Ours) & 74.42\%                                                                & 42.73\%                                                                & \emph{97.65\%}                           & 77.09\%                                                                & 13.48\%                           \\
                             & HMCAM (Ours)   & \textbf{74.92\%}                                      & \textbf{42.53\%}                                      & \emph{97.75\%}                           & \textbf{78.37\%}                                      & \textbf{14.11\%} \\ \hline\hline
\multirow{5}{*}{Densenet121} & FGSM           & 26.87\%                                                                & 29.40\%                                                                & 20.42\%                                                                & \emph{30.96\%}                           & 4.42\%                            \\
                             & PGD         & 63.38\%                                                                & 35.70\%                                                                & 57.22\%                                                                & \emph{95.34\%}                           & 5.67\%                            \\
                             & M-PGD        & 66.07\%                                                                & 39.16\%                                                                & 59.48\%                                                                & \emph{\textbf{97.83\%}} & 8.33\%                            \\
                             & AI-FGSM (Ours) & 69.64\%                                                                & 41.41\%                                                                & 63.35\%                                                                & \emph{96.49\%}                           & 9.77\%                            \\
                             & HMCAM (Ours)   & \textbf{69.82\%}                                      & \textbf{42.45\%}                                      & \textbf{63.87\%}                                      & \emph{96.39\%}                           & \textbf{10.36\%}
\end{tabular}
\caption{The success rates of several of non-targeted attacks against \textbf{a single network} on CIFAR10. The maximum perturbation is $\varepsilon=2/255$. The \emph{italic} columns in each block indicate white-box attacks while the rest are all black-box attacks which are more practical but challenging. Results have shown that our proposed methods (AI-FGSM and HMCAM) greatly improve the transferability of generated adversarial examples. We compare our AI-FGSM and HMCAM with FGSM, PGD and M-PGD (MI-FGSM+PGD), respectively.}\label{tab:single}
\end{table*}
\section{Experiment}
In this section, we conduct extensive experimental evaluations of our proposed methods on three benchmarks: CIFAR10\cite{krizhevsky2009learning}, ImageNet\cite{deng2009imagenet} and MNIST\cite{lecun1998mnist}. 
Firstly, we briefly introduce the major implementation settings in Section. \ref{sec_settings}, and perform comprehensive comparisons to verify the superiority of our HMCAM method on single and ensemble models in both white-box and black-box manners in Section. \ref{sec_Single} and Section. \ref{sec_Ensemble}.
Then, we perform detailed ablation studies to demonstrate the influence of different aspects in HMCAM and explore the possibility of few sample learning for competitive results in adversarial training in Section. \ref{sec_ablations}.
To further test the efficiency of CAT method in adversarial training, we provide detailed quantitative comparison results of our proposed models in Section. \ref{sec_CAT}.
Finally, to investigate the generalization of our approach, we also perform experiments on ImageNet against the champion solution in the defense track of CAAD 2018 competitions in Section. \ref{sec_competitions} and attempt to launch attack on public face recognition systems in Section. \ref{sec_online}. 

\subsection{Datasets and Implementation Details}\label{sec_settings}
\noindent \textbf{Datasets.} We employ the following four benchmark datasets for a comprehensive evaluation to validate the effectiveness of our HMCAM and CAT methods.
\begin{itemize}
  \item \textbf{CIFAR10}\cite{krizhevsky2009learning} is a widely used dataset consisting of 60,000 colour images of 10 categories. Each category has 6,000 images. Due to the resource limitation, we mainly focus on the CIFAR10\cite{krizhevsky2009learning} dataset with extensive experiments to validate the effectiveness of the proposed methods on both adversarial attack and training. 
  \item \textbf{ImageNet}\cite{deng2009imagenet} a large dataset with 1,283,166 images in the training set and 50,000 images in the validation set images collected from the Web. It has 1,000 synsets used to label the images. As it is extremely time-consuming to train a model from scratch on ImageNet, we only use it to test the generalization of our approach, which fights against the champion solution in the defense track of CAAD 2018 competitions.
  \item \textbf{MNIST}\cite{lecun1998mnist} is a database for handwritten digit classification. It consists of 60,000 training images and 10,000 test images, which are all $28\times 28$ greyscale images, representing the digits 0$\sim$9. In this experiment, we only perform different adversarial training methods on MNIST.

\end{itemize}

\noindent \textbf{Implementation details.} For adversarial attack, we pick six models, including four normally trained single models (ResNet32\cite{he2016deep}, VGG16 (without BN)\cite{simonyan2014very}, ResNetXt29-8-64\cite{xie2017aggregated} and DenseNet121\cite{huang2017densely}) and one adversarially trained ensemble models ($\operatorname{Resnet32}_{A}$). The hyper-parameters of different attack methods follow the default settings in \cite{art2018} and the total iteration number is set to $N=100$ (in most cases $T=N$ except HMCAM). We fix $T=50$ and $S=2$ for HMCAM, and the decay rate $\mu$ is set to 1.0 for M-PGD (MI-FGSM+PGD). The magnitude of maximum perturbation at each pixel is $\varepsilon=2/255$. For simplicity, we only report the results based on $L_{\infty}$ norm for the non-targetd attack.

For adversarial training, we follow the training scheme used in Free\cite{shafahi2019adversarial} and YOPO\cite{zhang2019you} on CIFAR10. We choose the standard Wide ResNet-34 and Preact-ResNet18 following previous works\cite{DBLP:conf/iclr/MadryMSTV18,zhang2019you}. For PGD adversarial training, we set the total epoch number $N=105$ as a common practice. The initial learning rate is set to 5e-2, reduced by 10 times at epoch 79, 90 and 100. We use a batch size of 256, a weight decay of 5e-4 and a momentum of 0.9 for both algorithms. During evaluating, we test the robustness of the model under CW\cite{carlini2017towards}, M-PGD and 20 steps of PGD with step size $\varepsilon=2/255$ and magnitude of perturbation $\varepsilon=8/255$ based on $L_{\infty}$ norm. When performing YOPO and Free, we train the models for 40 epochs and the initial learning rate is set to 0.2, reduced by 10 times at epoch 30 and 36. As for ImageNet, we fix the total loop times $T*K=4$ same as Free-4\cite{shafahi2019adversarial} for fair comparison. For all methods, we use a batch size of 256, and SGD optimizer with momentum 0.9 and a weight decay of 1e-4. The initial learning rate is 0.1 and the learning rate is decayed by 10 every $30/TK$ epochs. We also set step size $\epsilon=4/255$ and magnitude of perturbation $\varepsilon=4/255$ based on $L_{\infty}$ norm.

\subsection{Attacking a Single Model}\label{sec_Single}
We compare the attack success rates of HMCAM with the family of FGSM on a single network in Table \ref{tab:single}. The adversarial examples are created by one of the six networks in turns and test on all of them. The \emph{italic} columns in each block indicate white-box attacks and others refer to black-box attacks. From the Table \ref{tab:single}, we can observe that HMCAM outperforms all other FGSM family attacks by a large margin in black-box scenario, and maintains comparable results on all white-box attacks with M-PGD. 
For example, HMCAM obtains success rates of 74.92\% on ResNetXt29-8-64 (white-box attack), 78.37\% on DenseNet121 (black-box attack on normally trained model) and 14.11\% on $\operatorname{Resnet32}_{A}$ (black-box attack on adversarially trained model) if adversarial examples are crafted on ResNetXt29-8-64, while M-PGD only reaches the corresponding success rates of 72.81\%, 42.53\% and 10.11\%, respectively.
Considering that the white-box attack is usually used as a launch pad for the black-box attack, this demonstrates the practicality and effectiveness of our HMCAM for improving the transferability of adversarial examples. 

Note that AI-FGSM is a special case of HMCAM ($T=N$, $S=1$), which means AI-FGSM only carries out the inner loop in Algorithm \ref{alg:ours} for position and momentum updating. But AI-FGSM also reaches much higher success rates than FSGM family. This shows the superiority of our accumulated momentum strategy. 

\subsection{Attacking an Ensemble of Models}\label{sec_Ensemble}
Although our AI-FGSM and HMCAM better improve the success rates for attacking model in black-box scenario, the results of all the attack methods on adversarially trained model, e.g., $\operatorname{Resnet32}_{A}$, are far from satisfactory. To solve this problem, generating adversarial examples on the ensemble models\cite{DBLP:conf/iclr/LiuCLS17,dong2018boosting,xie2019improving} rather than a single model have been broadly adopted in the black-box scenario for enhancing the transferability and shown its effectiveness.

For the ensemble-based strategy, each one of the six models introduced above will be selected as the hold-out model while the rest build up an ensemble model. The ensemble weights are set equally for all the six models. The results are shown in Table \ref{tab:ensemble}. The \emph{ensemble} block consists of the \emph{white-box} attack which uses the ensemble model to attack itself, and the \emph{hold-out} block is composed of the \emph{black-box} attack that utilizes the ensemble model to generate adversarial examples for its corresponding hold-out model. 

We can observe from Table \ref{tab:ensemble} that our AI-FGSM and HMCAM always show much better transferability than other methods no matter which target model is selected. For example, the adversarial examples generated by an ensemble of ResNet32, VGG16 and DenseNet121 (ResNetXt29-8-64 hold-out) can fool ResNetXt29-8-64 with a 83.07\% success rate. Moreover, our proposed methods can remarkably boost the transferability of adversarial examples on adversarially trained model.
\renewcommand{\arraystretch}{1.2}
\begin{table*}[!htb]
\centering
\scriptsize
\begin{tabular}{l|cc|cc|cc|cc|cc}
\multicolumn{1}{c|}{\multirow{2}{*}{Attack}} & \multicolumn{2}{c|}{-ResNet32}      & \multicolumn{2}{c|}{-VGG16}         & \multicolumn{2}{c|}{-ResNetXt29-8-64} & \multicolumn{2}{c|}{-DenseNet121}   & \multicolumn{2}{c}{-ResNet32\_A}    \\ \cline{2-11} 
\multicolumn{1}{c|}{}                        & Ensemble         & Hold-out         & Ensemble         & Hold-out         & Ensemble          & Hold-out          & Ensemble         & Hold-out         & Ensemble         & Hold-out         \\ \hline\hline
FGSM                                         & 28.27\%          & 30.74\%          & 31.08\%          & 31.43\%          & 29.47\%           & 25.34\%           & 29.64\%          & 27.70\%          & 28.45\%          & 7.53\%           \\
PGD                                       & 90.97\%          & 80.73\%          & 94.79\%          & 50.30\%          & 91.05\%           & 80.14\%           & 92.13\%          & 82.11\%          & 92.11\%          & 13.88\%          \\
M-PGD                                      & 92.13\%          & 81.92\%          & \textbf{96.47\%} & 52.48\%          & 92.60\%           & 80.83\%           & 93.60\%          & 83.85\%          & 92.48\%          & 21.23\%          \\
AI-FGSM (Ours)                               & 92.62\%          & 83.12\%          & 96.06\%          & 56.54\%          & 92.94\%           & 82.29\%           & 93.47\%          & 84.46\%          & 93.04\%          & 28.31\%          \\
HMCAM (Ours)                                 & \textbf{92.81\%} & \textbf{83.76\%} & 96.29\%          & \textbf{57.43\%} & \textbf{92.99\%}  & \textbf{83.07\%}  & \textbf{93.88\%} & \textbf{85.38\%} & \textbf{94.18\%} & \textbf{30.11\%}
\end{tabular}
\caption{The success rates of several of non-targeted attacks against \textbf{an ensemble of networks} on CIFAR10. The maximum perturbation is $\varepsilon=2/255$. We report the results on the ensemble network itself (white-box scenario) and its corresponding hold-out network (black-box scenario). Model with ``-'' indicates it is the hold-out network. We compare our AI-FGSM and HMCAM with FGSM, PGD and M-PGD (MI-FGSM+PGD), respectively.}\label{tab:ensemble}
\end{table*}

\subsection{Ablation Study on Adversarial Attack}\label{sec_ablations} 
In the following sections, we perform several ablation experiments to investigate how different aspects of HMCAM influence its effectiveness. For simplicity, we only attack five single models introduced in the previous section, and focus on comparing our HMCAM with M-PGD since M-PGD is one of the most effective iterative attack method so far. We report the results in both white-box and black-box scenarios.

\subsubsection{Influence of Iteration Number} 
\begin{figure}[!htbp]
\centering
\begin{minipage}[t]{0.22\textwidth}
\centering
\includegraphics[width=\textwidth]{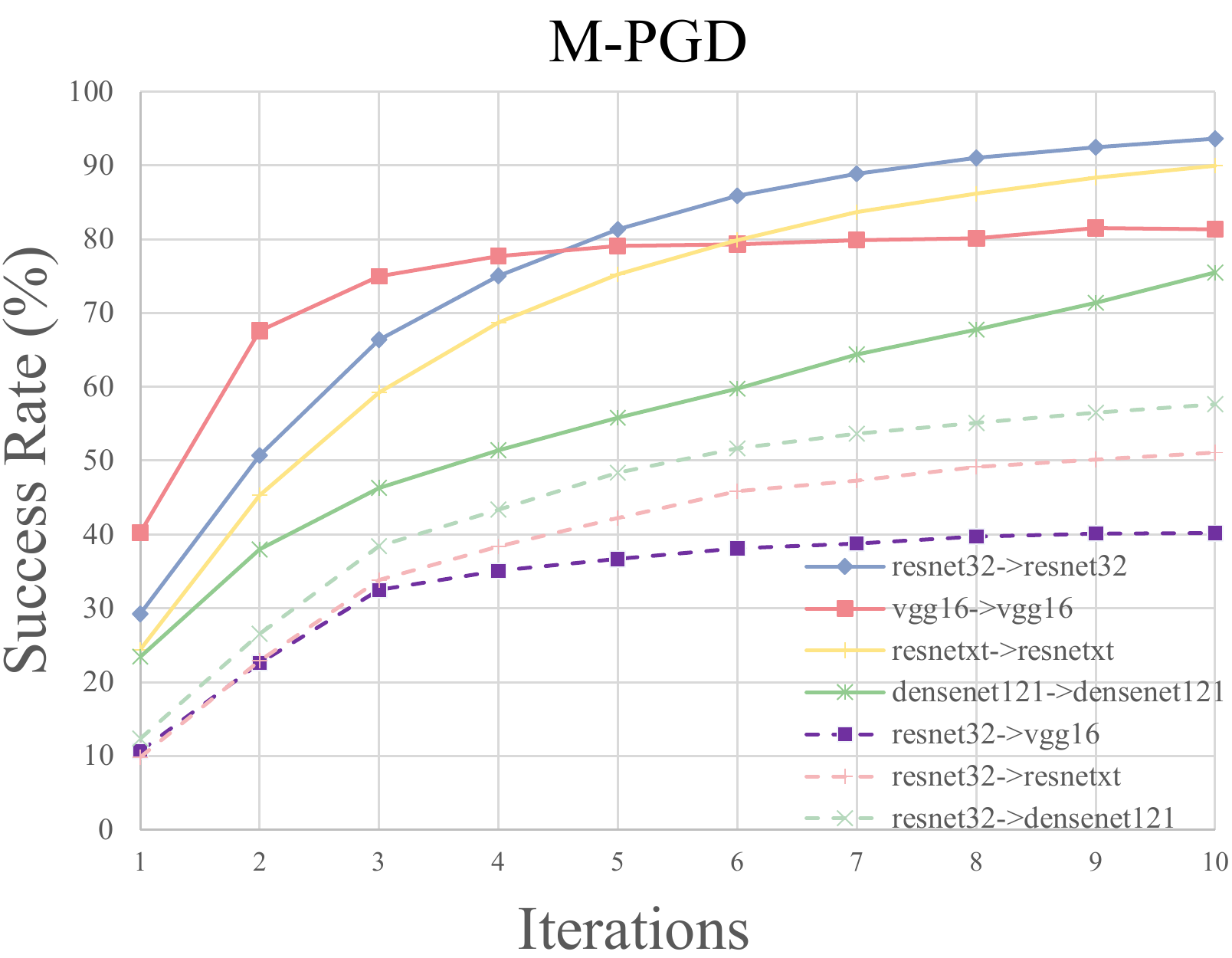}
\end{minipage}
\hspace{0.01\textwidth}
\begin{minipage}[t]{0.22\textwidth}
\centering
\includegraphics[width=\textwidth]{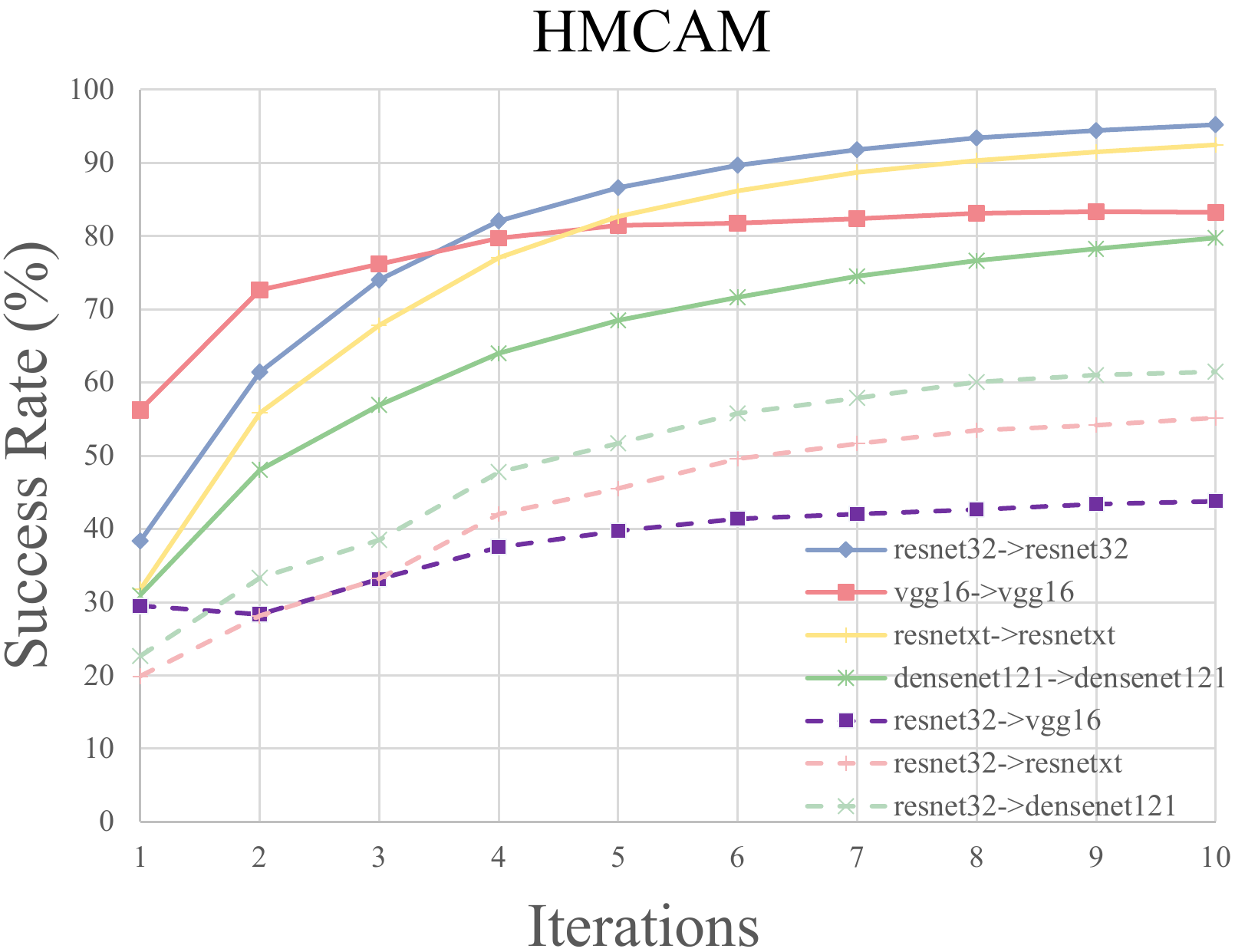}
\end{minipage}
\caption{The success rates of M-PGD (left) and HMCAM (right) on CIFAR10 over the first 10 iterations, with $\varepsilon=2/255$. Solid lines represent the white-box attacks and dashed lines represent the black-box attacks. ``$A\rightarrow B$'' means that model B is attacked by adversarial examples generated by model A.}\label{fig:iterations}
\end{figure}

To further demonstrate how fast our proposed method converges, we first study the influence of the total iteration number $N$ on the success rates. We clip a snippet over a time span of 10 iterations from the very beginning. Results are shown in Fig. \ref{fig:iterations}.

These results indicate that (1) the success rate of HMCAM against both white-box and black-box models are higher than M-PGD at all stages when combining with the extensive comparisons in Table \ref{tab:single}, which shows the strength of our HMCAM. (2) Even when the number of iterations is one order lower than that in Table \ref{tab:single}, the success rate of both HMCAM and M-PGD are still higher than PGD on the black-box scenario. Moreover, HMCAM ($N=10$) reaches higher values than PGD ($N=100$), demonstrating that HMCAM has strong attack ability and fast converges on both the white-box and black-box scenarios.

\renewcommand{\arraystretch}{1.0}
\begin{table*}[!htb]
\centering
\begin{tabular}{l|cccc|c}
Methods       & Natural                 & PGD-20 Attack           & M-PGD-20 Attack       & CW Attack               & Speed (mins) \\ \hline\hline
Natural train & 93.78\%                 & 0.00\%                  & 0.00\%                  & 0.00\%                  & 47                   \\ \hline\hline
PGD-10\cite{DBLP:conf/iclr/MadryMSTV18}        & \textbf{84.96\%$\pm$0.12\%} & 41.58\%$\pm$0.11\%          & 39.47\%$\pm$0.27\%          & 58.88\%$\pm$0.33\%          & 132                  \\
Free-8\cite{shafahi2019adversarial}           & 82.44\%$\pm$0.37\%          & 42.07\%$\pm$0.44\%          & 41.88\%$\pm$0.53\%          & 57.02\%$\pm$0.22\%          & 110                  \\
YOPO-5-3\cite{zhang2019you}      & 82.65\%$\pm$0.75\%          & 42.56\%$\pm$0.83\%          & 41.85\%$\pm$0.44\%          & 56.93\%$\pm$0.71\%          & \textbf{66}          \\ \hline\hline
CAT (Ours)    & 81.54\%$\pm$0.31\%          & \textbf{49.37\%$\pm$0.27\%} & \textbf{48.56\%$\pm$0.09\%} & \textbf{61.28\%$\pm$0.29\%} & 114                 
\end{tabular}
\caption{Validation accuracy and robustness of Preact-ResNet18 on CIFAR10. The maximum perturbation of all the attackers is $\varepsilon=8/255$. We report average over 5 runs on a single NVIDIA GeForce GTX XP GPU. The best result under different attack methods is in bold.}\label{tab:prenet}
\end{table*}
\renewcommand{\arraystretch}{1.0}
\begin{table*}[!htb]
\centering
\begin{tabular}{l|cccc|c}
Methods       & Natural               & PGD-20 Attack        & M-PGD-20 Attack     & CW Attack            & Speed (mins) \\ \hline\hline
Natural train & 94.58\%               & 0.00\%               & 0.00\%                & 0.00\%               & 212                  \\ \hline\hline
PGD-10\cite{DBLP:conf/iclr/MadryMSTV18}        & \textbf{87.11\%$\pm$0.37\%} & 48.4\%$\pm$0.22\%          & 44.37\%$\pm$0.11\%          & 45.91\%$\pm$0.14\%         & 2602                 \\
Free-8\cite{shafahi2019adversarial}        & 84.29\%$\pm$1.44\%          & 47.8\%$\pm$1.32\%          & 47.01\%$\pm$0.19\%          & 46.71\%$\pm$0.22\%         & 646                  \\
YOPO-5-3\cite{zhang2019you}      & 84.72\%$\pm$1.23\%          & 46.4\%$\pm$1.49\%          & 47.24\%$\pm$0.25\%          & 47.5\%$\pm$0.37\%          & \textbf{457}         \\ \hline\hline
CAT (Ours)    & 85.39\%$\pm$0.33\%          & \textbf{53.3\%$\pm$0.64\%} & \textbf{52.41\%$\pm$0.18\%} & \textbf{52.55\%$\pm$0.2\%} & 672                 
\end{tabular}
\caption{Validation accuracy and robustness of Wide ResNet34 on CIFAR10. The maximum perturbation of all the attackers is $\varepsilon=8/255$. We report average over 5 runs on a single NVIDIA GeForce GTX XP GPU. The best result under different attack methods is in bold.}\label{tab:wide34}
\end{table*}

\subsubsection{Influence of Step Size} 
\begin{figure}[!htbp]
\centering
\begin{minipage}[t]{0.22\textwidth}
\centering
\includegraphics[width=\textwidth]{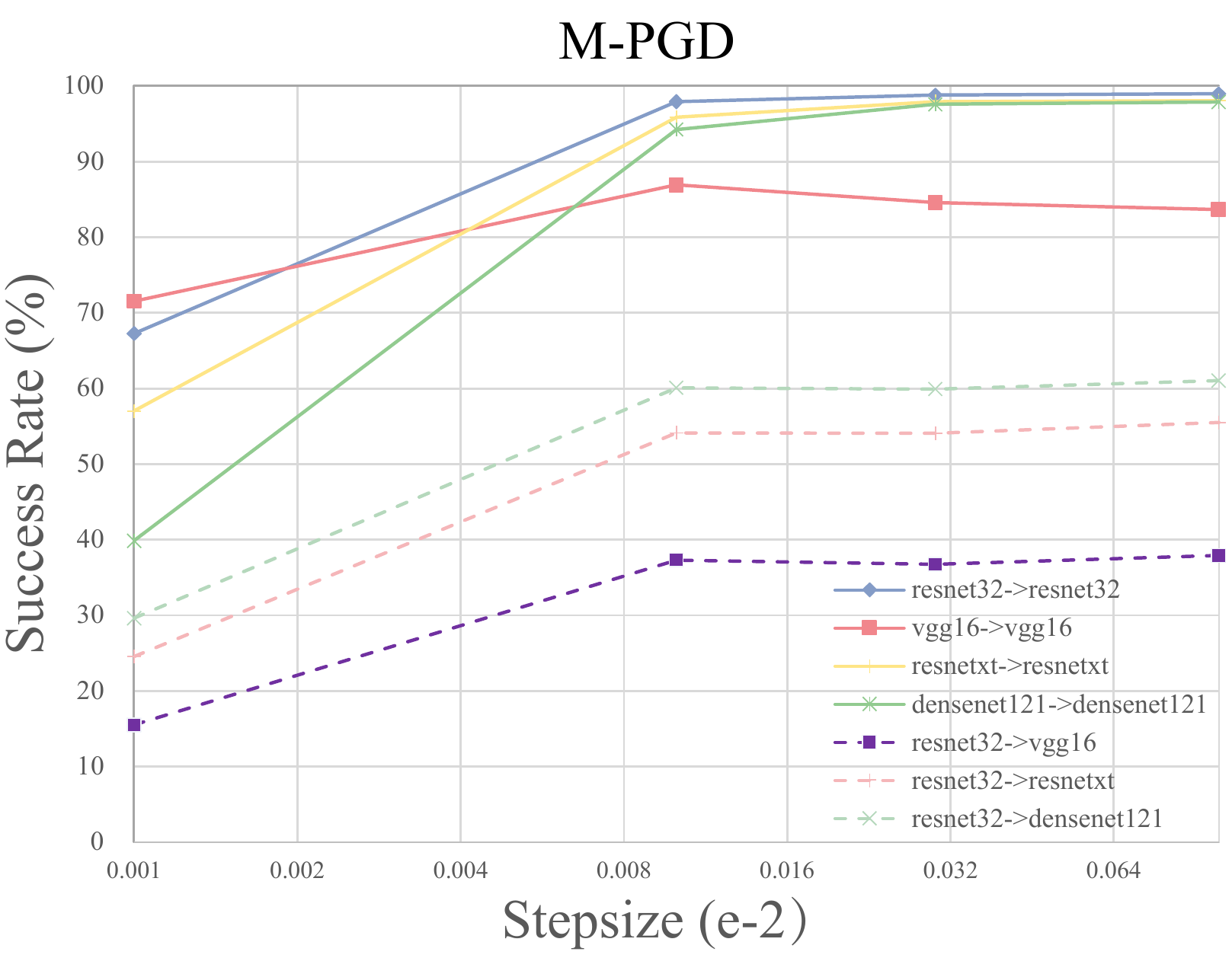}
\end{minipage}
\hspace{0.01\textwidth}
\begin{minipage}[t]{0.22\textwidth}
\centering
\includegraphics[width=\textwidth]{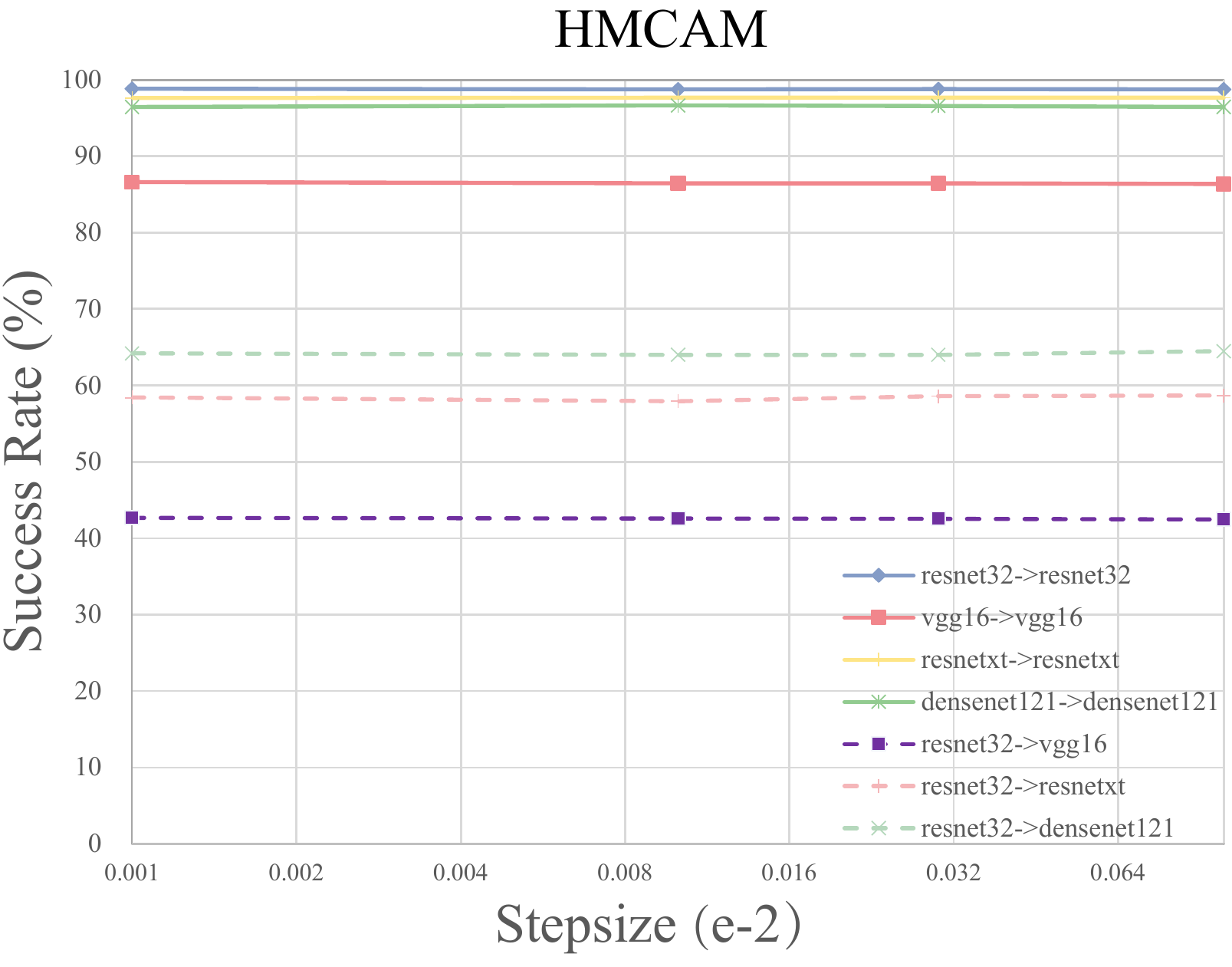}
\end{minipage}
\caption{The success rates of M-PGD (left) and HMCAM (right) on CIFAR10 after 100 iterations, with $\varepsilon=2/255$. Solid lines represent the white-box attacks and dashed lines represent the black-box attacks. ``$A\rightarrow B$'' means that model B is attacked by adversarial examples generating by model A.}\label{fig:stepsize}
\end{figure}

We also study the influence of the step size $\alpha$ on the success rates under both white-box and black-box settings. For simplicity, we fix the total iteration $N=100$ and set $S=1$ for HMCAM. We control the step size $\alpha$ in the range of $\left\{0.001, 0.01, 0.03, 0.1\right\}\times e^{-2}$. The results are plotted in Fig. \ref{fig:stepsize}. It can be observed that HMCAM outperforms M-PGD on both small and large step size. Under both the white-box and the black-box settings, our HMCAM is insensitive to the step size attributing to the accumulated momentum strategy.

\subsubsection{Fewer samples for competitive results} 
\begin{figure}[!htb]
\centering
\includegraphics[scale=0.45]{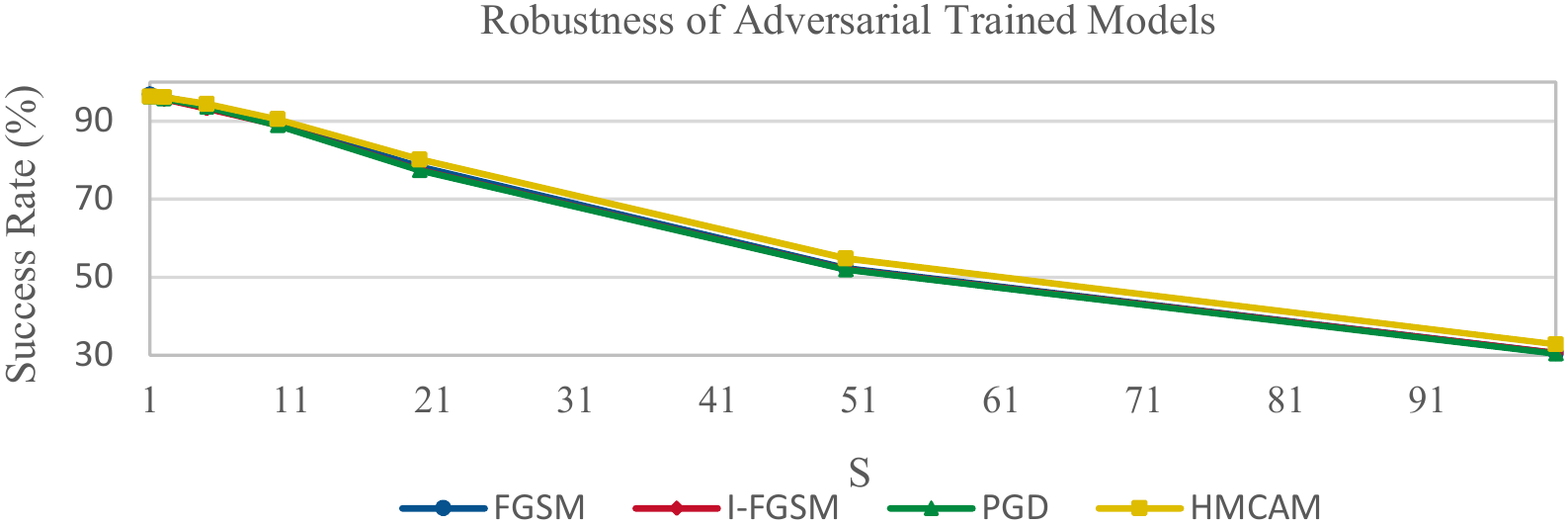}
\caption{Comparison with different adversarial training methods on CIFAR10. We use M-PGD as the attacker and report its success rate, with $\varepsilon=2/255$. Our HMCAM can use two orders of magnitude fewer samples than other methods to simulate the target distribution.}\label{fig:robustness}
\end{figure}

Since HMCAM is able to explore the distribution of adversarial examples, we finally investigate what aspects of systems are strengthened by our method. We also investigate whether the competitive result can be achieved with fewer samples when compared to the regular adversarial training. We generate adversarial images using FGSM, BIM and PGD to adversarially retrain the model and remain M-PGD to attack. We fix the total iteration $N=S*T=100$. To test the diversity of our generated samples, we select only $d=50$ samples from the whole training set for generating adversarial samples, then mixed into the training set for adversarial training. For fair comparison, we allow other methods except HMCAM to select more samples satisfying $d'=d*S$. We sweep the sampling number $S$ among $\left\{1,2,5,10,20,50,100\right\}$. The results are plotted in Fig. \ref{fig:robustness}. It is clear to see that the system trained by our HMCAM, only using two orders of magnitude fewer natural samples than any other method, can achieve comparable robustness. Considering the compared methods utilize the extra samples truly on the adversarial manifold, this indicates that our HMCAM draws the distribution of adversarial examples with few samples indeed.

\subsection{Efficiency for Adversarial Training}\label{sec_CAT}
In this subsection, we investigate whether the training time of adversarial training can benefit from the view of HMC since the high computational cost of adversarial training can be easily attributed to the long trajectory of MCMC finding the stationary distribution of adversarial examples. We take fixed but small number $k$ of transitions from the data sample as the initial values of the MCMC chains and then use these $k$-step MCMC samples to approximate the gradient for updating the parameters of model. We calculate the deviation value of the last 5 evaluations and report the average over 5 runs. Results about Preact-ResNet18 and Wide ResNet34 on CIFAR10 are shown in Table \ref{tab:prenet} and Table \ref{tab:wide34}, respectively. Our CAT method greatly boost the robust accuracy in a reasonable training speed. 

\begin{figure}[!htb]
\centering
\includegraphics[scale=0.45]{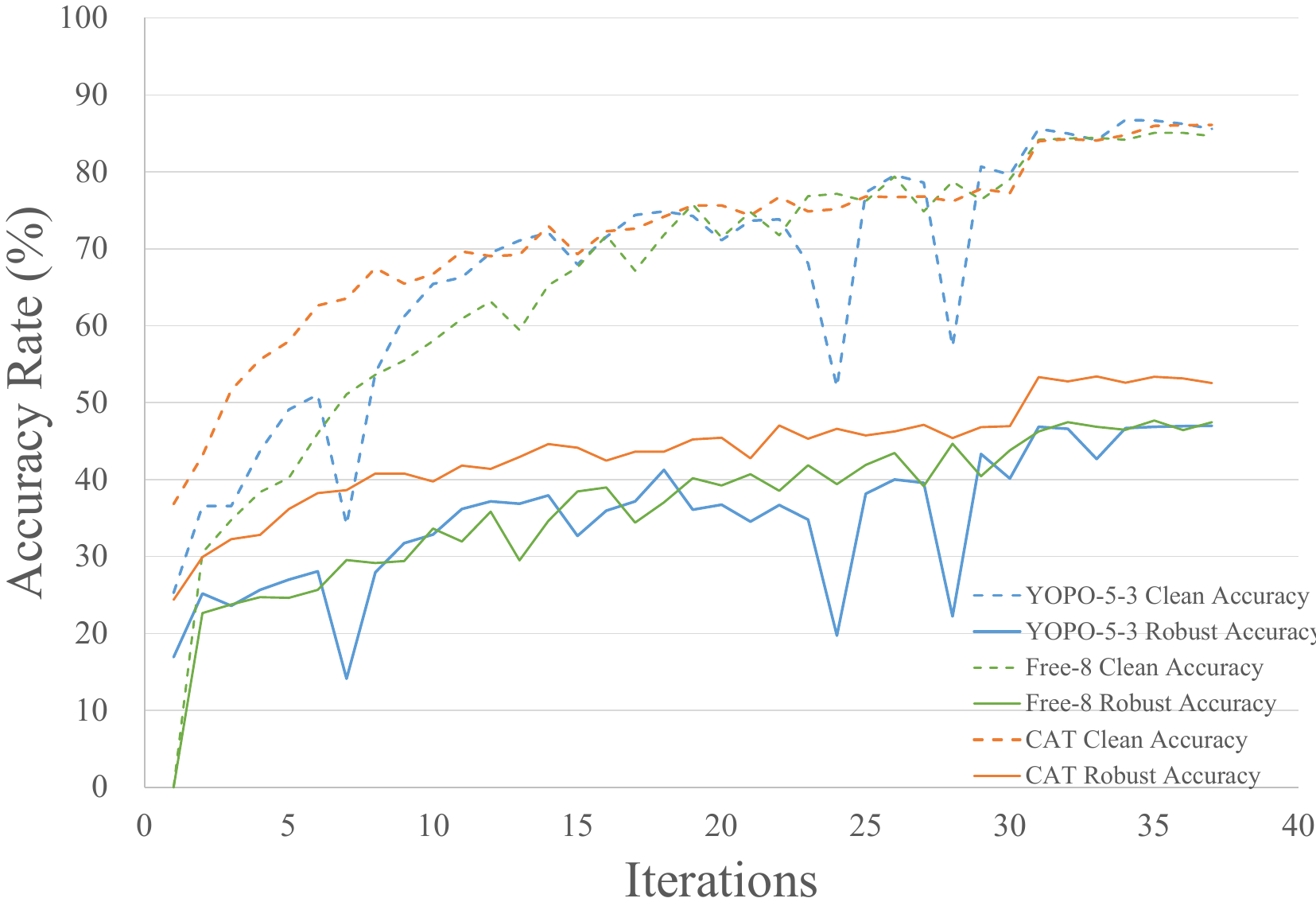}
\caption{Comparison with different adversarial training methods on both clean accuracy and robust accuracy (against PGD-10 with $\varepsilon=8/255$) of Wide ResNet34 on CIFAR10 at every iteration.}\label{fig:adv_training}
\end{figure}

We also present a comparison in terms of both clean accuracy and robust accuracy per iteration on all methods evaluated during training in Figure. \ref{fig:adv_training}. 
When compared with YOPO, the robust accuracy of our CAT method rises steadily and quickly while YOPO vibrates greatly and frequently. 

For ImageNet, we report the average results over last three runs. Comparison between free adversarial training and ours are shown in Table \ref{tab:ImageNet}. 
Although the 2-PGD trained ResNet-50 model still maintains its leading role in the best robust accuracy, it takes three times longer than our CAT method. Actually, when compared with its high computational cost of ImageNet training, this performance gain can be considered inefficient or even impractical for resource limited entities.
We also compare ResNet-50 model trained by our CAT method with the Free-4 trained, model trained by CAT produces much more robust models than Free-4 against different attacks in almost the same order of time.
\renewcommand{\arraystretch}{1.0}
\begin{table*}[!htb]
\centering
\begin{tabular}{l|ccccc|c}
Methods        & Clean Data                     & PGD-10 Attack                               & PGD-20 Attack                               & PGD-50 Attack                               & MI-FGSM-20 Attack       & \multicolumn{1}{l}{Speed (mins)} \\ \hline\hline
Natural train  & 75.34\%            & 0.14\%                               & 0.06\%                               & 0.03\%                               & 0.03\%           & 1437                             \\ \hline\hline
PGD            & \textbf{63.95\%} & \textbf{36.89\%} & \textbf{36.44\%} & \textbf{36.17\%} & \textbf{35.29\%} & 8928                             \\
Free-4         & 60.26\%                     & 31.12\%                              & 30.29\%                              & 30.07\%                              & 29.43\%          & \textbf{2745}                             \\ \hline\hline
CAT (Ours)     & 59.23\%                     & \textbf{35.91\%}                     & \textbf{35.72\%}                     & \textbf{35.76\%}                     & \textbf{34.67\%} & \textbf{2992}                            
\end{tabular}
\caption{Validation accuracy and robustness of ResNet50 on ImageNet. The maximum perturbation of all the attackers is $\varepsilon=4/255$. We report average over the final 3 runs. The best (or almost best) results under different attack methods are in bold. Our CAT achieves a trade-off between efficiency and accuracy.}\label{tab:ImageNet}
\end{table*}

We also investigate our CAT method on MNIST. We choose a simple ConvNet with four convolutional layers followed by three fully connected layers, which is of the same as \cite{zhang2019you}. For PGD adversarial training, we train the models for 55 epochs. The initial learning rate is set to 0.1, reduced by 10 times at epoch 45. We use a batch size of 256, a weight decay of 5e-4 and a momentum of 0.9. For evaluating, we perform a PGD-40 and CW attack against our model and set the size of perturbation as $\varepsilon=0.3$ based on $L_{\infty}$ norm as a common practice \cite{DBLP:conf/iclr/MadryMSTV18,zhang2019you,DBLP:conf/icml/ZhangYJXGJ19}. Results are shown in Table \ref{tab:small_net}.
\renewcommand{\arraystretch}{1.0}
\begin{table*}[!htb]
\centering
\begin{tabular}{l|ccc}
          & Clean Data       & PGD-40 Attack    & CW Attack        \\ \hline\hline
PGD-40    & 99.50\%          & 97.17\%          & 93.27\%          \\
Free-10   & 98.29\%          & 95.33\%          & 92.66\%          \\
YOFO-5-10 & \textbf{99.98\%} & 94.79\%          & 92.58\%          \\ \hline\hline
CAT (Ours)      & 99.36\%          & \textbf{97.48\%} & \textbf{94.77\%}
\end{tabular}
\caption{Validation accuracy and robustness of a small CNN on MNIST. The maximum perturbation of all the attackers is $\varepsilon=0.3$. The best result under different attack methods is in bold.}\label{tab:small_net}
\end{table*}

\subsection{Competitions and Real World Systems Attack}
\subsubsection{Attack CAAD 2018 Defense Champion}\label{sec_competitions}
Adversarial Attacks and Defenses (CAAD) 2018 is an open competition involving an exciting security challenge which stimulate the interest of a wide range of talents from industry and academia on adversarial learning. In the defense track of CAAD 2018, the champion solution\cite{xie2019feature} devised new network architectures with novel non-local means blocks and better adversarial training scheme, which greatly surpassed the runner-up approach under a strict criterion. We download the meticulously pretrained models\footnote{https://github.com/facebookresearch/ImageNet-Adversarial-Training/blob/master/INSTRUCTIONS.md} and apply our proposed method to attack the approach with default settings. We compare three attack methods (baseline, M-PGD and our HMCAM) on $\operatorname{ResNet152}_{A}$, $\operatorname{ResNet152}_{AD}$ and $\operatorname{ResNeXt101}_{AD}$ with 10/100 attack iterations, where $A$ is denoted as using adversarial training and $D$ presents being equipped with denoising blocks. Results are shown in Table \ref{tab:CAAD}. Note that the baseline attack method is one of the strongest white-box attacker as recent works\cite{xie2019feature,kannan2018adversarial}. From the Table \ref{tab:CAAD}, we can see that M-PGD is ineffective for attacking adversarially trained models with denoising blocks. Our proposed method outperforms both official baseline method and M-PGD. 
It is worth mentioning that our proposed method also outperforms one of the recent distributionally adversarial attack method DAA\cite{zheng2019distributionally}, which proposes a specific energy functionals combined the cross-entropy loss with the KL-divergence term for better adversarial-sample generation. Actually, DAA can be considered as a special case of the potential energy $U$.

\begin{table*}[!ht]
\centering
\begin{tabular}{l|ccc}
\multicolumn{1}{c|}{\multirow{2}{*}{Methods}} & \multicolumn{3}{c}{10/100-step Success Rate (\%)}                  \\ \cline{2-4} 
\multicolumn{1}{c|}{}                         & $\operatorname{ResNet152}_{A}$       & $\operatorname{ResNet152}_{AD}$ & $\operatorname{ResNeXt101}_{AD}$ \\ \hline\hline
PGD\cite{DBLP:conf/iclr/MadryMSTV18}                                      & 5.48/31.04           & 4.93/27.65          & 5.00/31.56            \\
M-PGD\cite{dong2018boosting}                & 4.01/24.63           & 3.51/22.07          & 3.44/23.78                   \\
DAA\cite{zheng2019distributionally}         & 4.23/27.31           & 4.17/24.69          & 4.55/28.07                             \\
HMCAM (Ours)                                 & \textbf{17.29/35.07} & \textbf{14.69/31.52}     & \textbf{17.54/36.36}
\end{tabular}
\caption{The success rates of targeted white-box attacks on ImageNet. The maximum perturbation is $\varepsilon=16/255$. We report three advanced adversarial attacks and our HMCAM on adversarially trained models with ($\operatorname{ResNet152}_{AD}$/$\operatorname{ResNeXt101}_{AD}$) and without ($\operatorname{ResNet152}_{A}$) feature denoising module.}\label{tab:CAAD}
\end{table*}

\subsubsection{Attack on Public Face Recognition Systems}\label{sec_online}
To further show the practical applicability of attack, we apply our HMCAM to the real-world celebrity recognition APIs in Clarifai\footnote{https://clarifai.com/models/celebrity-image-recognition-model-e466caa0619f444ab97497640cefc4dc}, AWS\footnote{https://aws.amazon.com/blogs/aws/amazon-rekognition-update-celebrity-recognition/} and Azure\footnote{https://azure.microsoft.com/en-us/services/cognitive-services/computer-vision/}. These celebrity recognition APIs allow users to upload any face images and recognize the identity of them with confidence score. The users have no knowledge about the dataset and types of models used behind these online systems. 
\begin{table}[!ht]
\centering
\footnotesize
\begin{tabular}{l|ccc}
\multicolumn{1}{c|}{\multirow{2}{*}{Methods}} & \multicolumn{3}{c}{Success Cases}         \\ \cline{2-4} 
\multicolumn{1}{c|}{}                         & Clarifai & AWS & Azure \\ \hline\hline
Geekpwn CAAD 2018                             & 3       & 0   & 0                         \\
$\mathcal{N}$Attack\cite{li2019nattack}      & 2       & 0   & 0                         \\
HMCAM (Ours)                                  & \textbf{8}       & \textbf{2}   & \textbf{1}                        
\end{tabular}
\caption{The results of our targeted attack on the real-world celebrity recognition APIs in Clarifai, AWS and Azure. We randomly selected 10 pairs of images and adopt a strict criterion called \emph{``all-or-nothing''} for our HMCAM attacker, which means that the success case counts only if \textbf{all} the adversarial examples in our generated sequence can fool the systems. The maximum perturbation is $\varepsilon=16/255$.}\label{tab:online}
\end{table}
We choose 10 pairs of images from the LFW dataset and learn perturbations from local facenet model to launch targeted attack, whose goal is to mislead the API to recognize the adversarial images as our selected identity. We randomly pick up 10 celebrities as victims from Google and 10 existing celebrities as targets from LFW, ensuring that all colors and genders are taken into account. Then we apply the same strategy as Geekpwn CAAD 2018 method that pulls victims towards their corresponding targets by the inner product of their feature vectors and generates noise to them. Finally, we examine their categories and confidence scores by uploading these adversarial examples to the online systems API.

We fix $\varepsilon=16/255$ and total iteration number $N=100$. Besides, we also set $S=5$ to generate a sequence of adversarial examples to test the robustness of these online systems.
Here we propose a \emph{strict evaluation criterion} derived from\cite{xie2019feature} for our HMCAM attacker, which we also call \emph{``all-or-nothing''}: an attack is considered successful only if \textbf{all} the adversarial examples in our generated sequence can deceive the system. This is a challenging evaluation scenario.
As shown in Table \ref{tab:online}, quite a part of them pass the recognition of the online systems and output the results we want. The qualitative results are given in 
the supplementary document. Note that we also compare our HMCAM method with one of state-of-the-art black-box attack method $\mathcal{N}$Attack\cite{li2019nattack}, which aims at finding a probability density distribution around the input and estimates the gradient by a modified NES\cite{wierstra2011natural} method. Comparisons between $\mathcal{N}$Attack and HMCAM show that the samples generated by our proposed method have the stronger transferability since HMCAM is just a white-box attack method.

\section{Conclusion}
In this paper, we formulate the generation of adversarial examples as a MCMC process and present an efficient paradigm called Hamiltonian Monte Carlo with Accumulated Momentum (HMCAM).
In contrast to traditional iterative attack methods that aim to generate a single optimal adversarial example in one run, HMCAM can efficiently explore the distribution space to search multiple solutions and generate a sequence of adversarial examples.
We also develop a new generative method called Contrastive Adversarial Training (CAT), which approaches equilibrium distribution of adversarial examples with only few iterations by building from small modifications of the standard Contrastive Divergence. 
Extensive results with comparisons on CIFAR10 showed that not only HMCAM attained much higher success rates than other black-box models and comparable results as other white-box models in adversarial attack, but also CAT achieved a trade-off between efficiency and accuracy in adversarial training.
By further evaluating this enhanced attack against the champion solution in the defense track of CAAD 2018 competition, HMCAM outperforms the official baseline attack and M-PGD.
To demonstrate its practical applicability, we apply the proposed HMCAM method to investigate the robustness of real-world celebrity recognition systems, and compare against the Geekpwn CAAD 2018 method.
The result shows that the existing real-world celebrity recognition systems are extremely vulnerable to adversarial attacks in the black-box scenario since most examples generated by our approach can mislead the system with high confidence, which raises security concerns for developing more robust celebrity recognition models. 
The proposed attack strategy leads to a new paradigm for generating adversarial examples, which can potentially assess the robustness of networks and inspire stronger adversarial learning methods in the future.








 
\bibliographystyle{IEEEtran}
\bibliography{mybibfile}

\clearpage
\twocolumn
\begin{IEEEbiography}[{\includegraphics[width=1in,height=1.25in,clip,keepaspectratio]{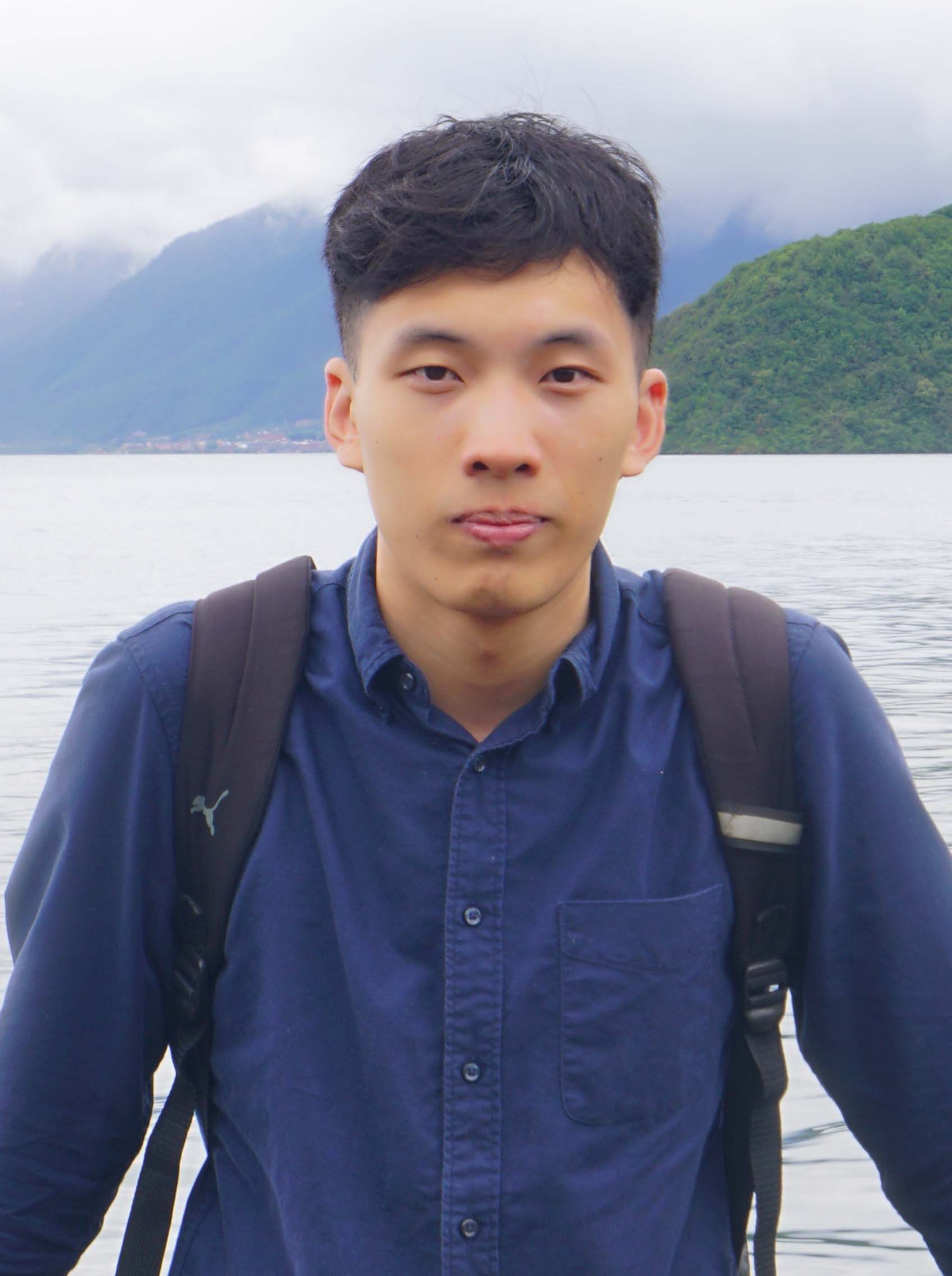}}]{Hongjun Wang} (S'20) received his B.E. degree of information security from Sun Yat-Sen University, Guangzhou, China, in 2018. He is currently working toward the M.E. degree at Sun Yat-Sen University. His current research interests include computer vision and the security of machine learning, particularly in adversarial attacks and defenses.
\end{IEEEbiography}

\begin{IEEEbiography}[{\includegraphics[width=1in,height=1.25in,clip,keepaspectratio]{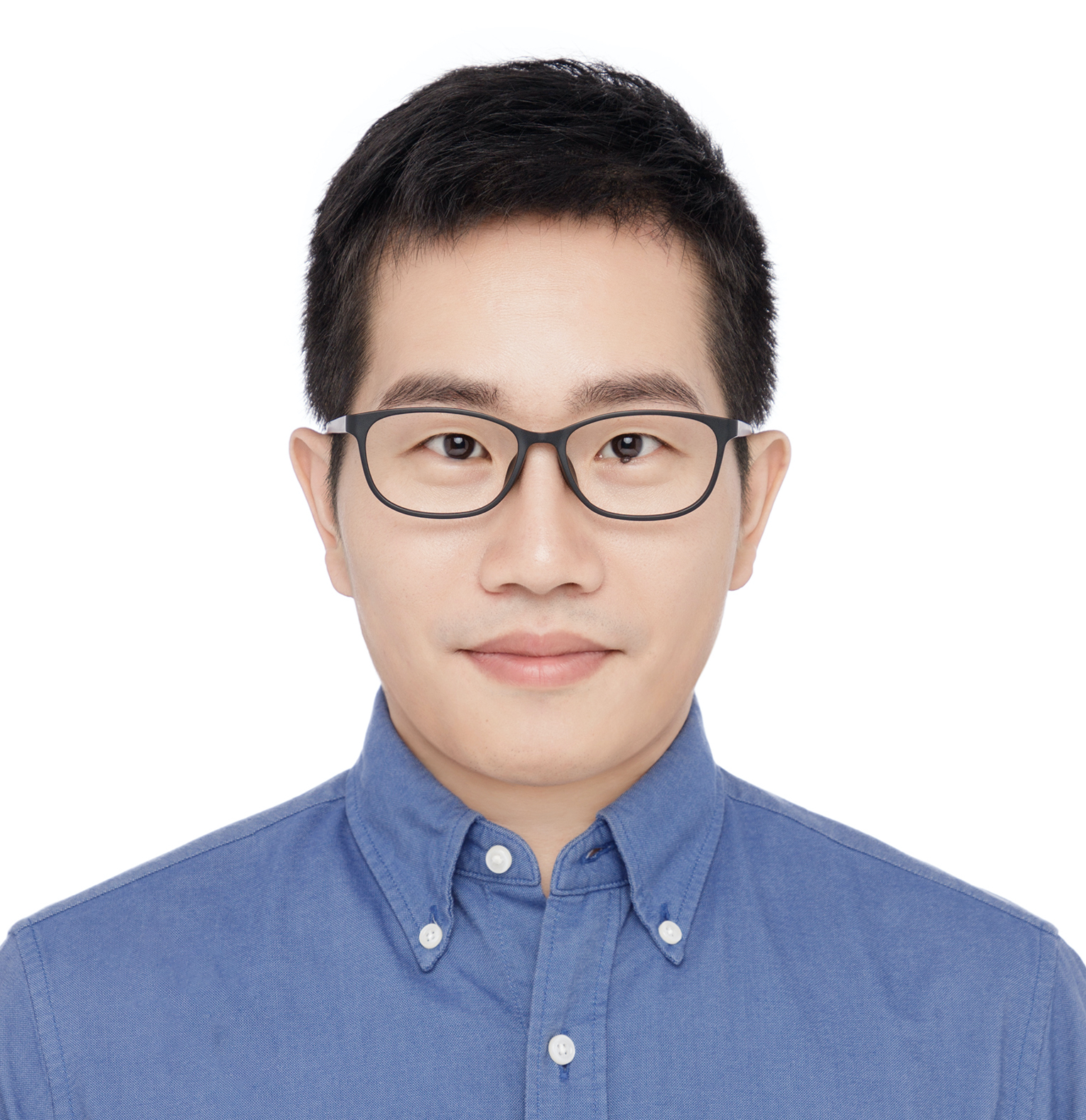}}]{Guanbin Li} (M'15) is currently an associate professor in School of Data and Computer Science, Sun Yat-sen University. He received his PhD degree from the University of Hong Kong in 2016. His current research interests include computer vision, image processing, and deep learning. He is a recipient of ICCV 2019 Best Paper Nomination Award. He has authorized and co-authorized on more than 60 papers in top-tier academic journals and conferences. He serves as an area chair for the conference of VISAPP. He has been serving as a reviewer for numerous academic journals and conferences such as TPAMI, IJCV, TIP, TMM, TCyb, CVPR, ICCV, ECCV and NeurIPS.
\end{IEEEbiography}

\begin{IEEEbiography}[{\includegraphics[width=1in,height=1.25in,clip,keepaspectratio]{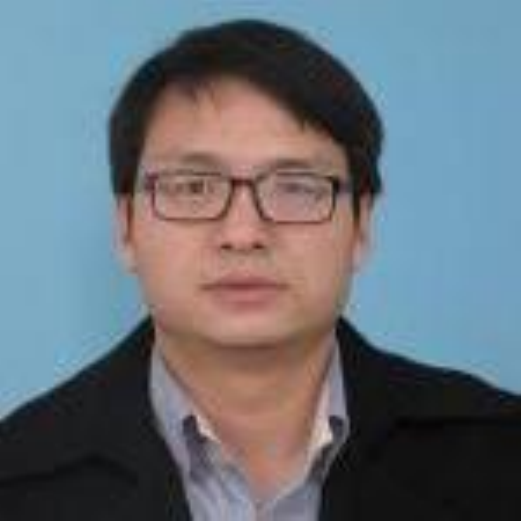}}]{Xiaobai Liu} is currently an Associate Professor of Computer Science in the San Diego State University (SDSU), San Diego. He received his PhD from the Huazhong University of Science and Technology, China. His research interests focus on scene parsing with a variety of topics, e.g. joint inference for recognition and reconstruction, commonsense reasoning, etc. He has published 60+ peer-reviewed articles in top-tier conferences (e.g. ICCV, CVPR etc.) and leading journals (e.g. TPAMI, TIP etc.). He received a number of awards for his academic contribution, including the 2013 outstanding thesis award by CCF(China Computer Federation). He is a member of IEEE.
\end{IEEEbiography}

\begin{IEEEbiography}[{\includegraphics[width=1in,height=1.25in,clip,keepaspectratio]{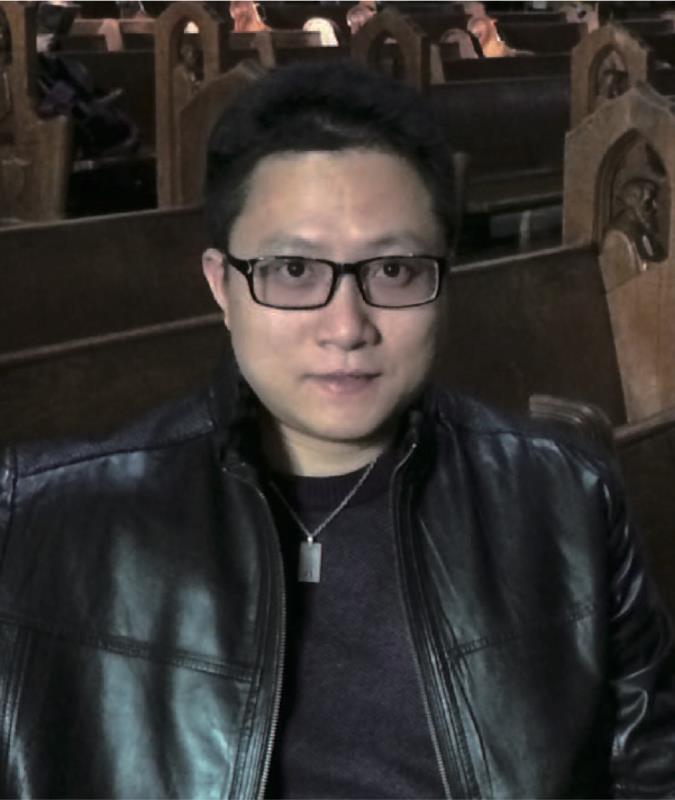}}]{Liang Lin} (M'09, SM'15) is a full Professor of Sun Yat-sen University. He is an Excellent Young Scientist of the National Natural Science Foundation of China. From 2008 to 2010, he was a Post-Doctoral Fellow at the University of California, Los Angeles. From 2014 to 2015, as a senior visiting scholar, he was with The Hong Kong Polytechnic University and The Chinese University of Hong Kong. He currently leads the SenseTime R$\&$D teams to develop cutting-edge and deliverable solutions on computer vision, data analysis and mining, and intelligent robotic systems. He has authored and co-authored more than 100 papers in top-tier academic journals and conferences. He has been serving as an associate editor of IEEE Trans. Human-Machine Systems, The Visual Computer and Neurocomputing. He served as area/session chairs for numerous conferences, such as ICME, ACCV, ICMR. He was the recipient of the Best Paper Runners-Up Award in ACM NPAR 2010, the Google Faculty Award in 2012, the Best Paper Diamond Award in IEEE ICME 2017, and the Hong Kong Scholars Award in 2014. He is a Fellow of IET.
\end{IEEEbiography}




\end{document}